\pdfoutput=1
\documentclass{article}
\usepackage[accepted]{icml2023}
\usepackage{microtype}
\usepackage{graphicx}
\usepackage{subfigure}
\usepackage{booktabs} 
\usepackage{enumitem}
\usepackage{algpseudocode}
\newcommand*\Let[2]{\State #1 $\gets$ #2}

\algrenewcommand\algorithmicrequire{\textbf{Input:}}
\algrenewcommand\algorithmicensure{\textbf{Output:}}
\usepackage{hyperref}

\usepackage{graphicx}
\usepackage{natbib}
\usepackage{doi}

\usepackage{colortbl} 
\definecolor{myblue}{RGB}{0, 128, 255}
\usepackage{amsmath}
\usepackage{cleveref}
\usepackage{amssymb}
\usepackage{mathtools}
\usepackage{amsthm}
\usepackage{array}
\usepackage{arydshln}
\usepackage{adjustbox}
\usepackage[super]{nth}
\usepackage{xspace}
\newcommand{\xhdr}[1]{\vspace{0em}\noindent{{\bf #1.}}}
\newcommand{\xhdrd}[1]{\vspace{0em}\noindent{{\bf #1}}}

\newcolumntype{R}[2]{%
    >{\adjustbox{angle=#1,lap=\width-(#2)}\bgroup}%
    l%
    <{\egroup}%
}
\newcommand{\methodname}{\textsc{Raincoat}\xspace}
\newcommand{\cmarkb}{\checkmark}
\newcommand{\source}{\mathcal{D}^s}
\newcommand{\target}{\mathcal{D}^t}
\newcommand{\be}{\mathbf{e}} 
\newcommand{\bw}{\mathbf{w}} 
\newcommand{\bx}{\mathbf{x}} 
\newcommand{\by}{y} 
\newcommand{\bz}{\mathbf{z}} 
\newcommand{\bv}{\mathbf{v}} 
\newcommand{\ba}{\mathbf{a}} 
\newcommand{\bp}{\mathbf{p}} 
\newcommand{\bmu}{\boldsymbol{\mu}} 
\newcommand{\bnu}{\boldsymbol{\nu}} 
\newcommand{\bC}{\mathbf{C}} %
\newcommand{\bW}{\mathbf{W}} 
\newcommand{\Loss}{\mathcal{L}} 
\newcommand{\Freq}{\textsc{F}} 
\newcommand{\Time}{\textsc{T}} 
\theoremstyle{plain}
\newtheorem{theorem}{Theorem}[section]

\theoremstyle{definition}

\newtheorem{problem}[theorem]{Problem}

\theoremstyle{remark}

\icmltitlerunning{Domain Adaptation for Time Series Under Feature and Label Shifts}
\begin{document}
\twocolumn[
\icmltitle{Domain Adaptation for Time Series Under Feature and Label Shifts}



\icmlsetsymbol{equal}{*}

\begin{icmlauthorlist}
\icmlauthor{Huan He}{hms}
\icmlauthor{Owen Queen}{hms}
\icmlauthor{Teddy Koker}{mit}
\icmlauthor{Consuelo Cuevas}{mit}
\icmlauthor{Theodoros Tsiligkaridis}{mit}
\icmlauthor{Marinka Zitnik}{hms}
\end{icmlauthorlist}

\icmlaffiliation{hms}{Department of Biomedical Informatics, Harvard University}
\icmlaffiliation{mit}{Artificial Intelligence Technology, MIT Lincoln Laboratory}

\icmlcorrespondingauthor{Huan He, Theodoros Tsiligkaridis, Marinka Zitnik}{huan\_he@hms.harvard.edu, ttsili@ll.mit.edu, marinka@hms.harvard.edu}

\icmlkeywords{Machine Learning, ICML}

\vskip 0.3in
]
\printAffiliationsAndNotice{} 

\begin{abstract}
Unsupervised domain adaptation (UDA) enables the transfer of models trained on source domains to unlabeled target domains. However, transferring complex time series models presents challenges due to the dynamic temporal structure variations across domains. This leads to feature shifts in the time and frequency representations. Additionally, the label distributions of tasks in the source and target domains can differ significantly, posing difficulties in addressing label shifts and recognizing labels unique to the target domain. Effectively transferring complex time series models remains a formidable problem. We present \methodname, the first model for both closed-set and universal domain adaptation on complex time series. \methodname addresses feature and label shifts by considering both temporal and frequency features, aligning them across domains, and correcting for misalignments to facilitate the detection of private labels. Additionally, \methodname improves transferability by identifying label shifts in target domains. Our experiments with 5 datasets and 13 state-of-the-art UDA methods demonstrate that \methodname can improve transfer learning performance by up to 16.33\% and can handle both closed-set and universal domain adaptation.
\end{abstract}

\section{Introduction}

\begin{figure}[t!]
    \centering
    \includegraphics[width=0.9\linewidth]{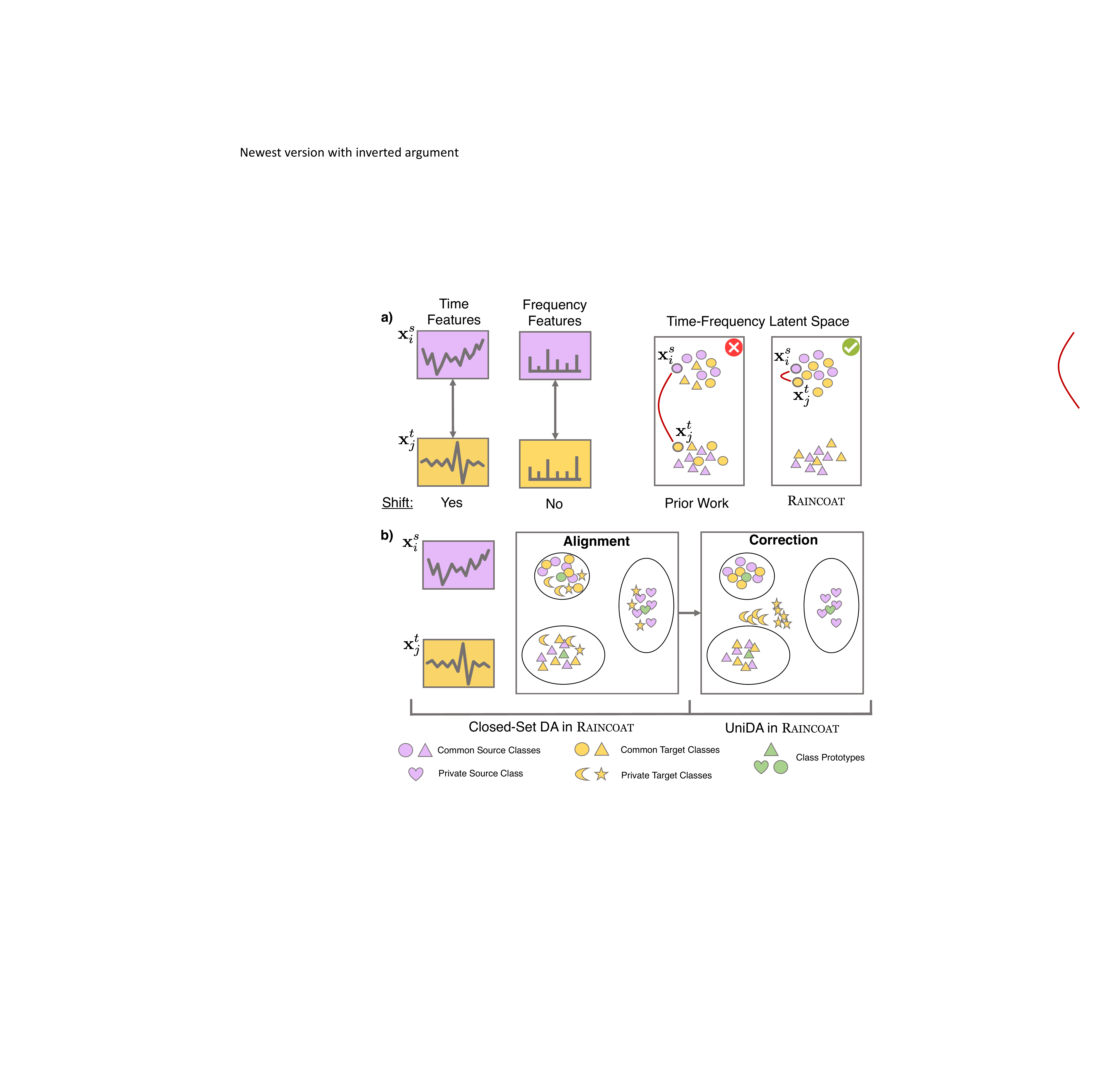}
    \caption{\textbf{a)} \methodname captures domain-invariant frequency features under feature and label shifts. \textbf{b)} For Closed-Set DA, \methodname aligns source and target domains for greater generalization. For Universal DA, the correction step prioritizes target-specific features to detect private target classes.}
    \label{fig:motivation}
\end{figure}





Neural networks have demonstrated impressive performance on time series datasets~\cite{ravuri2021skilful,lundberg2018explainable}. However, their performance deteriorates rapidly under domain shifts, making it challenging to deploy these models in real-world scenarios~\cite{TF-C,zhang2022graph}. Domain shifts occur when the test distribution is not identical to the training data, even though it is often related~\cite{wilds2021,Luo2018TakingAC,Zhang2013DomainAU}, meaning that latent representations do not generalize to test datasets drawn from different underlying distributions, even if the differences between these distributions are minor. To overcome these challenges, domain adaptation (DA) has emerged as a set of techniques that allow adaptation to new target domains and reduce bias by leveraging unlabeled data in target domains~\cite{Ganin:2016, DAN}.



Training models that can adapt to domain shifts is crucial for robust, real-world deployment. For instance, for healthcare time series, data collection methods vary widely across different clinical sites (domains)~\cite{zhang2022shifting}, leading to shifts in the underlying features and labels. It is preferable to train a model on a diverse dataset collected from multiple clinics rather than training and applying individual models on smaller, single-domain datasets for each clinic. Additionally, training a model that can detect unknown classes in test data, such as patients with rare diseases~\cite{alsentzer2022deep}, is advantageous for real-world implementation among end-users, such as clinicians~\cite{tonekaboni2019clinicians}. Endowing learning systems with DA capabilities can increase their reliability and expand applicability across downstream tasks.



DA is a highly complex problem due to several factors. First, models trained for robustness to domain shifts must learn highly generalizable features; however, neural networks trained using standard practices can rely on spurious correlations created by non-causal data artifacts~\cite{geirhos2020shortcut, degrave2021radiographic}, hindering their ability to transfer across domains. Additionally, shifts in label distributions across domains may result in \textit{private labels}, i.e., classes that exist in the target domain but not in the source domain \cite{Lipton2018DetectingAC}. In unsupervised DA, a model must generalize across domains when labels from the target domain are not available during training \cite{long2018conditional,Kang2019ContrastiveAN}. Therefore, DA methods must be able to identify when a private label is encountered in the target domain without any prior supervision on detecting these unknown labels \cite{You2019UniversalDA,Fu2020LearningTD}. Yet, that is not possible by techniques that rely on training samples that simulate predicting unknown labels. This highlights the need for time series DA methods that 1) produce \textit{generalizable representations robust to feature and label shifts}, and 2) expand the scope of existing DA methods by supporting both \textit{closed-set} and \textit{universal} DA.








DA becomes even more challenging when applied to time series data. Domain shifts can occur in both the time and frequency features of time series, which can create a shift that highly perturbs time features while frequency features are relatively unchanged, or vice versa (Figure~\ref{fig:motivation}a). Previous time series DA methods fail to explicitly model frequency features. Further, models can fail to generalize due to shortcut learning \cite{brown2022detecting}, which occurs when the model focuses on time-space features while overlooking crucial underlying concepts in the frequency-space domain, leading to limited poor performance on data unseen during training.
Additionally, universal DA---when no assumptions are made about the overlap between labels in the source and target domains---is an unexplored area in time series research (Figure~\ref{fig:motivation}b). 



\begin{figure*}
    \centering
    \includegraphics[width=0.7\linewidth]{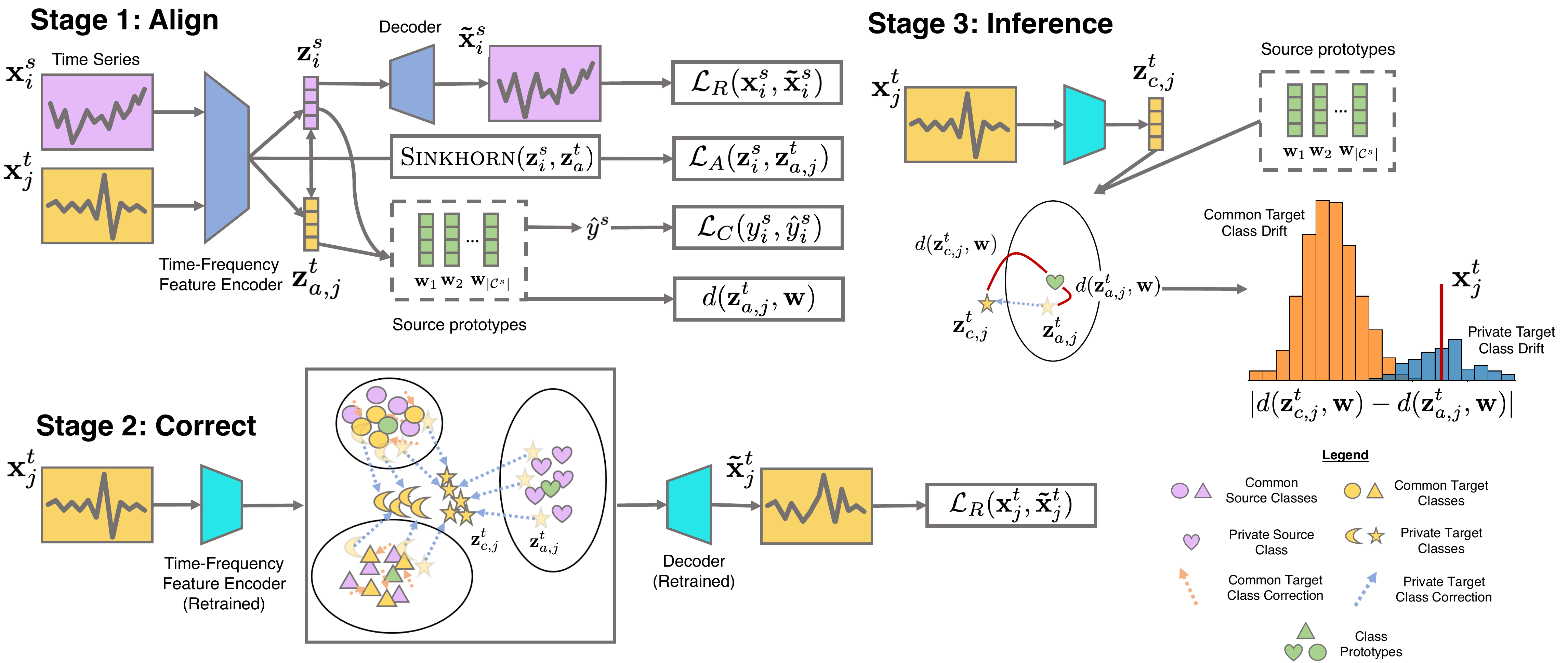}
    \caption{ Illustration of the \methodname method for time series DA. Details provided in-text.
}
    \label{fig:overview_framework}
\end{figure*}

\xhdr{Present Work}
We introduce \methodname (f\underline{R}equency-augmented \underline{A}l\underline{I}g\underline{N}-then-\underline{C}orrect for d\underline{O}main \underline{A}daptation for \underline{T}ime series), a novel domain adaptation method for time series data that can handle both feature and label shifts (as shown in Figure~\ref{fig:motivation}). {Our method is the first to address both closed-set and universal domain adaptation for time series and has the unique capability of handling feature and label shifts. To achieve this, we first use time and frequency-based encoders to learn time series representations, motivated by inductive bias that domain shifts can occur via both time or frequency feature shifts. We use Sinkhorn divergence for source-target feature alignment and provide both empirical evidence and theoretical justification for its superiority over other popular divergence measures. Finally, we introduce an ``align-then-correct" procedure for universal DA, which first aligns the source and target domains, retrains the encoder on the target domain to correct misalignments, and then measures the difference between the aligned and corrected representations of target samples to detect unknown target classes (as shown in Figure~\ref{fig:overview_framework}).} We evaluate \methodname on five time-series datasets from various modalities, including human activity recognition, mechanical fault detection, and electroencephalogram prediction. Our method outperforms strong baselines by up to 9.0\% for closed-set DA and 16.33\% for universal DA. \methodname is available at \url{https://github.com/mims-harvard/Raincoat}.

\section{Related Work} 
 
\xhdr{General Domain Adaptation}
General domain adaptation (DA), leveraging labeled source domain to predict labels on the unlabeled target domain, has a wide range of applications \cite{ganin2015unsupervised, sener2016learning, zhang2018task, perone2019unsupervised, ramponi2020neural}.  
We organize DA methods into three categories: 1) \textit{Adversarial training:} 
%
A domain discriminator is optimized to distinguish source and target domains,
while a deep classification model learns transferable features indistinguishable by the
domain discriminator \cite{Hoffman2015SimultaneousDT, tzeng2017adversarial,motiian2017few,long2018conditional,hoffman2018cycada}. 2) \textit{Statistical divergence:} These approaches aim to extract domain invariant features by minimizing domain discrepancy in a latent feature space. Widely used measures include MMD \cite{Rozantsev2016BeyondSW}, correlation alignment (CORAL) \cite{dcoral}, contrastive domain
discrepancy (CDD) \cite{Kang2019ContrastiveAN}, optimal transport distance \cite{courty2017joint,redko2019optimal}, and graph matching loss \cite{Yan2016ASS,Das2018UnsupervisedDA}. 3) \textit{Self-supervision:} These general DA approaches incorporate auxiliary self-supervision training tasks. These methods learn domain-invariant features through a pretext learning task, such as data augmentation and reconstruction, for which a target objective can be computed without supervision \cite{kang2019contrastive,singh2021clda,tang2021gradient}. In addition, reconstruction-based methods achieve alignment by carrying out source domain classification and reconstruction of target domain data or both source and target domain data \cite{ghifary2016deep, jhuo2012robust}. \methodname sits in the category of both 2 and 3. 

\xhdr{Domain Adaptation for Time Series}
While in light of successes in computer vision, limited methods have focused on adaptation approaches for time series data. To date, few DA methods are specifically designed for time series. 1) \textit{Adversarial training:} VRADA \cite{Purushotham2017VariationalRA} builds upon a variational recurrent neural network (VRNN) and trains adversarially to capture complex temporal relationships that are domain-invariant. CoDATS \cite{Wilson2020MultiSourceDD} builds upon VRADA but uses a convolutional neural network for the feature extractor. 2) \textit{Statistical divergence:} SASA~\cite{Cai2021TimeSD} aligns the condition distribution of the time series data by minimizing the discrepancy of the associative structure of time series variables between domains. AdvSKM~\cite{ijcai2021p378} and \cite{Ott2022DomainAF} are metric-based methods that align two domains by considering statistic divergence. 
3) \textit{Self-supervision:} DAF~\cite{DAF-icml} extracts domain-invariant and domain-specific features to perform forecasts for source and target domains through a shared attention module with a reconstruction task. CLUDA~\cite{Ozyurt2022ContrastiveLF} and CLADA~\cite{CALDA} are two contrastive DA methods that use augmentations to extract domain invariant and contextual features for prediction. 
However, the above methods align features without considering the potential gap between labels from both domains. Moreover, they focus on aligning only time features while ignoring the implicit frequency feature shift (Fig.~\ref{fig:motivation}a). In contrast, \methodname considers the frequency feature shift to mitigate both feature and label shift in DA. 

\xhdr{Universal Domain Adaptation}
Prevailing DA methods assume all labels in the target domain are also available in the source domain. This assumption, known as closed-set DA, posits that the domain gap is driven by feature shift (as opposed to label shift). However, the label overlap between the two domains is unknown in practice. Thus, assuming both feature and label shifts can cause the domain gap is more practical. In contrast to closed-set DA, universal domain adaptation (UniDA) \cite{You2019UniversalDA} can account for label shift. UniDA categorizes target samples into common labels (present in both source and target domains) or private labels (present in the target domain only). UAN \cite{You2019UniversalDA}, CMU \cite{Fu2020LearningTD}, and TNT \cite{Chen2022EvidentialNC} use sample-level uncertainty criteria to measure domain transferability. Samples with lower uncertainty are preferentially selected for adversarial adaptation. However, most UniDA methods detect common samples using sample-level criteria, requiring users to specify the threshold to recognize private labels. Moreover, over-reliance on source supervision neglects discriminative representation in the target domain. DANCE \cite{Saito2020UniversalDA} uses self-supervised neighborhood
clustering to learn features to discriminate private labels. Similarly, DCC \cite{Li2021DomainCC} enumerates cluster numbers of the target domain to obtain optimal cross-domain consensus clusters as common classes. Still, the consensus clusters are not robust enough due to challenging cluster assignments. MATHS \cite{Chen2022MutualNN} detects private labels via mutual nearest-neighbor contrastive learning. In contrast, UniOT \cite{Chang2022UnifiedOT} uses optimal transport to detect common samples and produce representations for samples in the target domain. 
However, these methods use a feature encoder shared across both domains even though the source and target domains are shifted. In addition, most require fine-tuned thresholds to recognize private labels. 

\section{Problem Setup and Formulation}
\label{sec:problem}
\xhdr{Notation} 
We are given a dataset $\mathcal{D} = \{(\bx_i, \by_i)\}_{i=1}^{n}$ of $n$ multivariate time series samples where $i$-th sample $\bx_i \in \mathbb{R}^{T \times d}$ contains readouts of $d$ sensors over $T$ time points. Without loss of generality, we consider regular time series --- \methodname can be used with techniques, such as Raindrop~\cite{zhang2022graph} to handle irregular time series. We use $\bx_i$ to denote a time series (both univariate and multivariate). 
Each label $\by_i$ in $\mathcal{D}$ belongs to the label set $\mathcal{C}$, i.e., $\by_i \in \mathcal{C}$. We use $\source=\left\{\left(\bx_i^s, \by_i^s\right)\right\}_{i=1}^{n_s}$ to denote
the source domain dataset with $n_s$ labeled samples, where $\bx_i^s$ is a source domain sample and $\by^s_i$ is the  associated label. The target domain dataset is unlabeled and denoted as $\target=\left\{\left(\bx_i^t\right)\right\}_{i=1}^{n_t}$ with $n_t$
unlabeled samples. Source and target label sets are denoted as $\mathcal{C}^s$ and $\mathcal{C}^t$, respectively. Zero, one or more labels may be shared between source and target domains, which we denote as $\mathcal{C}^{s,t} = \mathcal{C}^s \cap \mathcal{C}^t$. Source and target domains have samples drawn from source and target distributions, $\source \sim p_s(\bx^s, \by^s)$ and $\target \sim p_t(\bx^t, \by^t)$.





We consider two types of domain shifts: feature shift and label shift. Feature shift occurs when marginal probability distributions of $\bx$ differ, $p_s(\bx) \neq p_t(\bx)$, while conditional probability distributions remain constant across domains, $p_s(\by|\bx) = p_t(\by|\bx)$ \cite{Zhang2013DomainAU}. Label shift occurs when marginal probability distributions of $\by$ differ, $p_s(\by) \neq p_t(\by)$. Feature shifts may occur in time series due to, for example, differences in sensor measurement setup or length of samples. A unique property of time series is that feature shifts may occur in both time and frequency spectra. The importance of modeling shifts in both the time and frequency spectrum is discussed in later sections. Label shift may occur as either a change in the proportion of classes in either domain or as a categorical shift: both domains might contain different classes in their label sets.


\begin{problem}[\textbf{Closed-set Domain Adaptation for Time Series Classification}]
    Given the source and target domain time series datasets, $\source$ and $\target$, whose label sets are the same, $\mathcal{C}^s = \mathcal{C}^t$, and target labels $y^t$ are not available at train time. \methodname specifies a strategy to train a classifier $f$ on $\mathcal{D}^s$ such that $f$ generalizes to $\mathcal{D}^t$, \textit{i.e.}, it minimizes classification risk on $\mathcal{D}^t$: $\mathbb{E}_{\bx_i, \by_i \sim \target}\left[\Loss_{C}(f(\bx_i), y_i)\right]$, where $\Loss_C$ is a classification loss function.
    \label{prob:closedset}
\end{problem}

 In a real-world application, little information may be available on the feature or label distribution of the target domain. Private labels in either the source or target domain may exist, i.e., classes present in one domain but absent in the other. Thus, it is desirable to relax the strict assumption of $\mathcal{C}^s = \mathcal{C}^t$ made by Problem \ref{prob:closedset}. We denote source private labels as $\bar{\mathcal{C}}^s = \mathcal{C}^s \setminus \mathcal{C}^t$, target private labels as $\bar{\mathcal{C}}^t = \mathcal{C}^t \setminus \mathcal{C}^s$, and labels shared between domains as $\mathcal{C}^{s,t} = \mathcal{C}^s \cap \mathcal{C}^t$. We denote the access of samples in dataset $\mathcal{D}$ belonging to label set $\mathcal{C}$ as $\mathcal{D}[\mathcal{C}]$, e.g., samples in the target domain belonging to the common label set would be denoted as $\mathcal{D}^t[\mathcal{C}^{s,t}]$. Domains might not have common labels,  $\mathcal{C}^{s,t} = \emptyset$, leading to the definition of universal DA.

\begin{problem}[\textbf{Universal Domain Adaptation (UniDA) for Time Series Classification}]
Given our source and target domain time series datasets, $\source$ and $\target$, where target labels $y^t$ are unavailable at train time. \methodname specifies a stratefy to train a classifier $f$ on $\source$ such that $f$ generalizes to $\target$, \textit{i.e.}, it minimizes classification risk of a loss function $\Loss_C$ on samples belonging to $\mathcal{C}^{s,t}$ in $\target$: $\mathbb{E}_{\bx_i, \by_i \sim \target[\mathcal{C}^{s,t}]}\left[\Loss_{C}(f(\bx_i), y_i)\right]$, while identifying samples in \emph{private} target classes, $\bx_i \sim \mathcal{D}^t[\bar{\mathcal{C}}^t]$, as \emph{unknown} samples.


\label{prob:unida}
\end{problem}

\section{Preliminaries}
\label{sec:pre}
\xhdr{Discrete Fourier Transform}
%
Given a  series sample $\bx$ with $d$ channels and $T$ time points, it is transformed to the frequency space by applying the 1-dim DFT of length $T$ to each channel and then transforming it back using the 1-dim inverse DFT, defined as:
\vspace{-0.2cm}
\begin{equation}
    \begin{aligned}
    \label{eqn:DFT}
     \text{Forward DFT}:& \quad \bv[m]=\textstyle\sum_{t=0}^{T-1} \bx[t] \cdot e^{-i 2 \pi \frac{mt}{T}} \\
     \text{Inverse DFT}:& \quad
     \bx[n]=\frac{1}{T} \textstyle\sum_{t=0}^{T-1} \bv[m] \cdot e^{i \cdot 2 \pi \frac{mt}{T}}
    \end{aligned}
\end{equation}
where $T=$ number of points, $n$ = current point index, $m$ = current frequency index, where $m \in [0, T-1]$. 
We denote the extracted amplitude and phase as $\mathbf{a}$ and $\mathbf{p}$ respectively:
\vspace{-0.2cm}
\begin{equation}
\label{eqn:ap}
\begin{aligned}
    \mathbf{a}[m] &=\frac{\left|\bv[m]\right|}{T}=\frac{\sqrt{\operatorname{Re}\left(\bv[m]\right)^2+\operatorname{Im}\left(\bv[m]\right)^2}}{T} \\
    \mathbf{p}[m] &=\operatorname{atan2}\left(\operatorname{Im}\left(\bv[m]\right), \operatorname{Re}\left(\bv[m]\right)\right)
\end{aligned}
\end{equation}
where Im$(\bv[m])$ and Re$(\bv[m])$ indicate imaginary and real parts of a complex number, and atan2 is the two-argument form of arctan.

\section{\methodname Approach}

We start with an overview of \methodname and proceed with
 (5.2) time-frequency encoding, (5.3) feature alignment, (5.4) unknown sample detection, and (5.5) training and inference. 

\subsection{Overview}

\methodname is an unsupervised method for closed set and universal domain adaptation in time series, addressing Problems~\ref{prob:closedset}-\ref{prob:unida}. \methodname consists of three modules: a time-frequency encoder $G_{\textsc{T}\textsc{F}}$, a classifier $H$, and an auxiliary decoder $U_{\textsc{T}\textsc{F}}$. Sec.~\ref{sec:tf-encoder} describes the encoder $G_{\textsc{T}\textsc{F}}$, which leverages both time and frequency features. Sec.~\ref{sec:align} describes how Sinkhorn divergence is a suitable divergence measurement to align the source and target domain because frequency features may not share the same support across both domains. Sec.~\ref{sec: correct} motivates the correction step for UniDA. Sec.~\ref{sec:inference} describes how \methodname detects potential unknown samples through analysis of pre- and post-correction embeddings. Finally, Sec.~\ref{sec:training} provides an overview of \methodname models.





\subsection{Time-Frequency Feature Encoder}
\label{sec:tf-encoder}


We begin by highlighting the significance of frequency features in DA for time series. Although various methods have been proposed to solve the time series DA problem under the assumption of feature shift, none of them explicitly address situations where changes in the frequency domain also act as an implicit feature shift. To fill this gap, \methodname encodes both time and frequency features in its latent representations. The source frequency and time features are denoted as $\be_{\Freq, i}^s$ and $\be_{\Time, i}^s$, respectively, while the target frequency and time features are represented as $\be_{\Freq, i}^t$ and $\be_{\Time, i}^t$. For simplicity, the superscript indicating the source or target domain is omitted in the rest of the text.


\xhdr{Shift of Frequency Features}
We formalize the frequency shift of time series as another type of feature shift. For this purpose, we use the Fourier transform, with the possibility of exploring other options such as wavelets left for future work. A time series $\bx_i$ can be represented as a combination of sinusoids, each with a specific frequency, amplitude, and phase, as explained in Sec.~\ref{sec:pre}. If the conditional distributions of the labels with respect to the frequency features are equal ($p_s(y|DFT(\bx^s)) = p_t(y|DFT(\bx^t))$), but the domains have different frequency features ($p(DFT(\bx^s)) \neq p(DFT(\bx^t))$), then a frequency shift occurs.

\begin{figure}[t]
    \centering
 \includegraphics[width=0.75\linewidth]{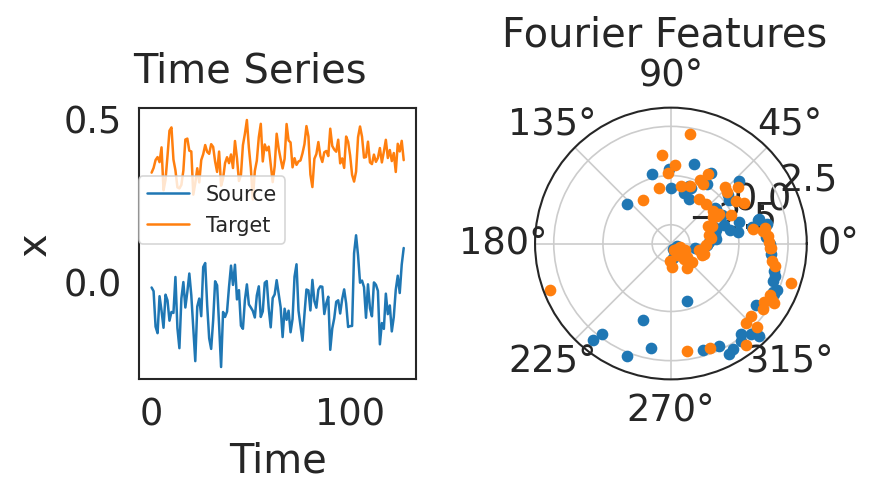}
    \caption{\textit{Left:} averaged sensor readings (one channel) of the walking activity collected from two persons (source and target). \textit{Right:}  corresponding polar coordinates of Fourier features.  Fourier features are more domain-invariant than time features.}
    \label{fig:fft}
\end{figure}

\xhdr{Frequency Features Promote Domain Adaptation}
\citeauthor{BenDavid2006AnalysisOR, BenDavid2010ATO} demonstrated that the performance of DA techniques is bounded by the divergence between the source and target domains, and that a small feature shift is necessary for DA techniques to be effective. {However, unsupervised DA methods for time series align only time features ($\be_{\Time, i}^s$ and $\be_{\Time, i}^t$), leading to sub-optimal performance when the time feature shift is large. By including frequency features in the encoder $G_{\Time\Freq}$, we can uncover potential invariant features across domains and improve transferability.} For instance, Figure \ref{fig:fft} illustrates the sensor readings of walking activity from two different individuals ($\bx_i^s$ and $\bx_i^t$) in the WISDM dataset \cite{10.1145/1964897.1964918} and their corresponding Fourier features ($\be_{\Freq, i}^s$ and $\be_{\Freq, i}^t$). Using only time features would result in poor predictions in the target domain due to a significant time feature shift between $\bx_i^s$ and $\bx_i^t$. On the other hand, frequency features from different domains do not exhibit significant feature shifts and thus are domain invariant. {This suggests that incorporating frequency features can lead to more accurate predictions in the target domain as DA aims to extract domain-invariant features.} For this reason, \methodname uses both time and frequency features in domain alignment.

\xhdr{Frequency Feature Encoder}
Inspired by Fourier neural operator (FNO)~\cite{li2021fourier}, \methodname applies convolution on low-frequency modes of the Fourier transform of $\bx_i$. We make two modifications to improve the utility of Fourier convolution for DA: \textit{1) Prevent Frequency Leakage:} Discrete Fourier Transform considers inputs $\bx_i$ to be periodic. Violation of such assumption results in frequency leakage \cite{Harris1978OnTU}. 
Specifically, given two window sliced time series $\bx_i^s$ and $\bx_i^t$ , applying DFT~ \eqref{eqn:DFT} could return perturbed and noisy $\bv_i^s$ and $\bv_i^t$ which may lead to noisy-biased domain alignment. To prevent aligning on noisy frequency features, \methodname applies a smoothing function (cosine function) before applying DFT. 
\textit{2) Consider amplitude $\ba_i$ and phase $\bp_i$ information:} Instead of using inverse DFT to convert $\bv_i$ back to time-space which is an unnecessary step for frequency feature extraction, \methodname extracts the polar coordinates of frequency coefficients to keep both low-level ($\ba_i$) and high-level ($\bp_i$) semantics. The frequency space features $\be_{\Freq}$ is a concatenation  $[\mathbf{a}_i;\mathbf{p}_i]$.

Now we summarize how $G_{\Freq\Time}$ encodes time-frequency feature from $\bx_i$. 
Define a convolution operator ``$\ast$" and weight matrix $\mathbf{B}$, the encoder $G_{\Freq}$ encodes frequency features $\be_{\Freq,i}$ by: \textit{1) Smooth:} $\bx_i =\operatorname{Smooth}(\bx_i)$,~\textit{2) DFT:} $\bv_i =\operatorname{DFT}(\bx_i)$,~\textit{3) Convolution:}
$\tilde{\bv}_{i}= \mathbf{B} \ast \bv_i$, \textit{4) Transform:} $\ba_i, \bp_i \leftarrow \tilde{\bv}_{i} \text{ (Use Eq.~\ref{eqn:ap})}$, \textit{5) Extract:} $ \be_{\Freq, i} = [\ba_i;\bp_i] $
The time features $\be_{\Time,i}$ can be obtained using any existing time feature encoder, such as CNNs. Finally, the latent representation $\bz_i$ is a concatenation of frequency and time features $[\be_{\Freq,i};\be_{\Time,i}]$. Details are in Appendix \ref{asec:encoder}.

\subsection{Domain Alignment of Time-Frequency Features}
\label{sec:align}
Next, we address the question of what is the appropriate metric to align frequency features between $\be_{\Freq}^s$ and $\be_{\Freq}^t$. \methodname represents the frequency features as the amplitude and phase, $\be_{\Freq,i}^s = [\mathbf{a}^s_i;\mathbf{p}^t_i],\be_{\Freq,i}^t = [\mathbf{a}^t_i;\mathbf{p}^t_i]$, meaning that the frequency feature shift can be represented as $p_s(\mathbf{a}^s, \mathbf{p}^s) \neq p_t(\mathbf{a}^t, \mathbf{p}^t)$.

\xhdr{Disjoint Support Sets for Frequency Features}
An appropriate metric to align frequency features between $\be_{\Freq}^s$ and $\be_{\Freq}^t$ is challenging to find. Distance measures such as the total variation distance or Kullback-Leibler divergence are not suitable because they are unstable when the supports of distributions are deformed and do not metricize the convergence in law~\citep{Feydy:2019}, meaning that they do not effectively capture the discrepancy when $\be^s_{\Freq,i}$ and $\be^t_{\Freq,i}$ have disjoint support. The KL divergence, for example, grows unbounded ($KL(\be_{\Freq,i}^s||\be_{\Freq,i}^t) \to +\infty$) when $\be_{\Freq,i}^t$ and $\be_{\Freq,i}^t$ are far apart, leading to a degradation of alignment and early collapse. An ideal  divergence measure could capture the discrepancy even if $\be^s_{\Freq,i}$ and $\be^t_{\Freq,i}$ have disjoint support ($\textrm{supp}(\be^s_{\Freq,i}) \cap \textrm{supp}(\be^t_{\Freq,i}) \approx \emptyset$).

The components of frequency features, amplitude $\ba$ and phase $\bp$, have different distributions. The phase $\bp$ has a uniform distribution over the range of polar angles, which makes it easy to measure the distance between $\mathbf{p}^s_i$ and $\mathbf{p}^t_i$, bounded in the polar coordinate system $\mathbf{p}_i \in [0, 2\pi)$. {However, the amplitude $\ba$ has a Rayleigh distribution with an unlimited scale, $\mathbf{a}_i \in [0, +\infty)$, making it difficult to measure the distance between $\mathbf{a}^s_i$ and $\mathbf{a}^t_i$ using the KL divergence. The KL divergence can not provide useful gradients when $\ba_i^s$ are $\ba_i^t$ are far apart. This leads to a lack of alignment when the amplitudes are far apart, as numerically verified in Figure \ref{fig:sinkhorn_motivation} in the Appendix.}


\xhdr{Sinkhorn Divergence}
The Sinkhorn divergence is an entropy-regularized optimal transport distance that enables the comparison of distributions with disjoint supports. Another metric, maximum mean discrepancy (MMD), addresses the issue of disjoint support by considering the geometry of the distributions. However, we demonstrate that MMD has a theoretical weakness that manifests as vanishing gradients or similar artifacts. To address this, \methodname aligns the source features ($\bz^s_i$) and target features ($\bz^t_i$) by minimizing a domain alignment loss based on Sinkhorn. Further details are provided in Appendix \ref{sec:appendixMMD}.


\subsection{Correction Step in \methodname}
\label{sec: correct}


In this section, we explain how the correction step helps reduce negative transfer by rejecting target unknown samples $\bx^t \sim \mathcal{D}^t[\bar{\mathcal{C}}^t]$. The correction step updates the encoder $G_{\Time\Freq}$ and decoder $U_{\Time\Freq}$ by solving a reconstruction task on target samples $\bx^t \sim \target$. This updated $G_{\Time\Freq}$ repositions the target features $\bz^t_i$. The target features before and after the correction step are denoted as $\bz^t_{a,i}$ and $\bz^t_{c,i}$, respectively.

\xhdr{Motivation for Reconstructing $\bx^t_i$}
The cluster assumption~\cite{Chapelle2005SemiSupervisedCB} holds that the input data is separated into clusters and that samples within the same cluster have the same label. {Based on this, we argue that preserving target discriminative features $\bz^t_i$ is important for UniDA, because such features help generate discriminative clusters, including clusters of target unknown samples, which improves UniDA.} To do this, \methodname minimizes a reconstruction loss to adapt the feature encoder $G_{\Time\Freq}$ and decoder $U_{\Time\Freq}$. The target features $\bz_{a,i}^t$ before the correction step are generated by a shared encoder $G_{\Time\Freq}$ that aligns the source and target domains. As a result, the target features of common samples $\bx^t \sim \mathcal{D}^t[\mathcal{C}^{s,t}]$ should change less in the latent space than those of target unknown samples $\bx^t \sim \mathcal{D}^t[\bar{\mathcal{C}}^t]$. {This indicates that the corrected encoder $G_{\Time\Freq}$ maintains the features of common target samples close to their originally assigned label while letting the features of target unknown samples diverge from their originally assigned label.} \methodname leverages this to detect and reject target unknown samples, which we discuss next.


\subsection{Inference: Detect Target Private Samples}
\label{sec:inference}
\methodname detects target unknown samples $\bx^t \sim \mathcal{D}^t[\bar{\mathcal{C}}^t]$ by determining the movement of target features before and after the correction step. It assumes that when the target domain contains unknown labels, the distribution of the movement will exhibit a bimodal structure.

For brevity, the feature vector $\bz^t_i$ is used as an input to $H$, which consists of prototypes for each class $\bW = [\bw_1, \bw_2, \cdots, \bw_{C}]$. Denote the distance (cosine similarity) of $\bz^t_i$ to its assigned prototype $c$ as $d(\bz^t_i,\bw_c)$. Cosine similarity is a reasonable choice because the cross entropy (CE) loss encourages angular separation. It can be interpreted as aligning the feature vectors $\bz^t_i$ along its assigned class prototype. The cosine similarity in the form of the dot product gives CE an intrinsic angular property, which is observed in Eq.~\ref{eqn: CE}
where features naturally separate in the polar coordinates with CE only. 
Given a target feature $\bz^t_i$ and true label $y_i=c$, the cross entropy can be expressed as:  \vspace{-0.5cm}
\begin{equation}
\label{eqn: CE}
\begin{array}{l}
\!\!\!\mathcal{L}_{\mathrm{CE}}(\hat{y}, y)=-\log \frac{\exp \left(\mathbf{w}_c^T\bz^t_i \right)}{\sum_j \exp \left(\mathbf{w}_j^T\bz^t_i\right)}
\propto \sum\limits_{j \neq c} \exp \left(\mathbf{w}_j^T\bz^t_i-\mathbf{w}_c^T\bz^t_i\right) \\ \\
\propto \sum\limits_{j \neq c} \exp \left(\|\bz^t_i\|_2\left\|\mathbf{w}_j\right\|_2 \cos \left(\theta_j\right)-\|\bz^t_i\|_2\left\|\mathbf{w}_c\right\|_2 \cos \left(\theta_c\right)\right) \nonumber\\
\end{array}
\end{equation}
As a result, if the target feature $\bz^t_i$ is close to its prototypes, then $d(\bz^t_i,\bw_c)$ will be small, and vice versa. Then \methodname measures the movement by calculating the absolute difference of target features' distance to the assigned prototype before and after correction given by $d^{ac}_i = |d(\bz^t_{a,i},\bw_c)- d(\bz^t_{c,i},\bw_c)|$.
    
Next, \methodname detects if there are private target samples in each class by first running a bimodal test on each group of $\mathcal{C}^{s}$. If the bimodal test tells us $d^{ac}$ has two modes, it then trains a $2$-mean cluster to fit the distribution of $d^{ac}$. For each class, after we obtain the centroid $\mu_1, \mu_2$, where $\mu_1 < \mu_2$, \methodname takes $\mu_2$ as our threshold to reject unknown target samples. 
 

\begin{algorithm}[t!]
\footnotesize
\caption{Overview of \methodname }
\label{alg:raincoat}
\begin{algorithmic}[1]
\State \textbf{Input}: dataset $\mathcal{D}^s$ , $\mathcal{D}^t$ ; epochs $E_1, E_2$; time-frequency feature encoder, $G_{\Time\Freq}$, and decoder, $U_{\Time\Freq}$ (Alg. \ref{alg:freq-align}); prototype classifier $H$ \\
\vspace{-0.2cm}\hrulefill
\State \textit{Stage 1: Alignment} (introduced in \ref{sec:tf-encoder} , \ref{sec:align}))
    \For {$E_1$ epochs}
        \State Extract $\bz^s_i, \bz^t_i \gets G_{\Time\Freq}(\bx^s_i), G_{\Time\Freq}(\bx^t_i)$
        \State $\mathcal{L}_{A} \gets \textsc{Sinkhorn}(
        \bz^s_i, \bz^t_i)$ (Alg. \ref{alg:sinkhorn})
        \State $\mathcal{L}_{R} \gets |\bx^s_i - U_{\Time\Freq}(\bz^s_i)|$ 
        \State $\mathcal{L}_{C} \gets CE(y^s_i, H(\bz^s_i))$
        \State Update $U_{\Time\Freq}, G_{\Time\Freq}, H$ with $\nabla (\Loss_A + \Loss_R + \Loss_C)$
    \EndFor \\
\hrulefill
\State \textit{Stage 2: Correction} (introduced in \ref{sec: correct})
    \State Extract features: $\bz^t_{a,i} \gets G_{\Time\Freq}(\bx^t_i)$
    \State Distance to prototypes: $\mathbf{d}_{\text{align}} \gets d(\bz^t_{a,i}, H)$
    \For {$E_2$ epochs}
        \State $\mathcal{L}_{R} \gets |\bx^t_i - (U_{\Time\Freq} \circ G_{\Time\Freq})(\bx^t_i)|$
        \State Update $U_{\Time\Freq}, G_{\Time\Freq}$ with $\nabla \mathcal{L}_{R}$
    \EndFor
    \State Extract post-correction: $\bz^t_{c,i} \gets G_{\Time\Freq}(\bx^t_I)$
    \State Re-compute: $\mathbf{d}_{\text{correct}} \gets d(\bz^t_{c,i}, H)$ \\
\hrulefill
\State \textit{Stage 3: Inference} (introduced in \ref{sec:inference})
    \State $d^{ac}_i = |d(\bz^t_{a,i},\bw_c)- d(\bz^t_{c,i},\bw_c)|$
    \For {$c$ in $C^s$}
        
        \State $p \gets \text{Bimodal Test}$
        \If{$p<0.05$}
        \Comment{Bimodal structure detected}
            \State $\mu_c^{\text{common}}, \mu_c^{\text{unknown}} =\textproc{Cluster}(d^{ac}|{\hat{y}=c})$
        \EndIf
    \EndFor
\end{algorithmic}
\end{algorithm}

\subsection{Overview of \methodname Models}
\label{sec:training}

During alignment, \methodname  trains a classifier $H$ using labeled source dataset $\source$ and a feature encoder $G_{\Time\Freq}$ and decoder $U_{\Time\Freq}$  using both $\source$ and $\target$. At the same time, it aligns target features $\bz^t_i$ with source features $\bz^s_i$ using Sinkhorn divergence. 
The overall loss function in this step has three terms. First, the sinkhorn distance $\Loss_{A}(\bz^t_i, \bz^s_i)$ urges the target features $\bz^t_i$ to be aligned with source features $\bz^s_i$. Second, the reconstruction loss $\Loss_R(\bx^s_i,U_{\Time\Freq}(G_{\Time\Freq}(\bx^s_i)))$ promotes learning of semantic features of $\source$. Third, the classification loss $\Loss_C(H(G_{\Time\Freq}(\bx^s_i)), \by^s_i)$ guides the model to classify samples correctly. In summary, the  loss in this step is defined as $\Loss = \Loss_A + \Loss_R + \Loss_C$.

In this step, target common samples could be classified correctly, and target unknown samples will be misclassified because the $G_{\Time\Freq}$ aligns all samples without considering the label shift. 
The correction step in \methodname aims to correct such negative transfer (target unknown samples) by exploiting target-specific discriminative features by minimizing $\Loss_R(\bx^t_i,U_{\Time\Freq}(G_{\Time\Freq}(\bx^t_i)))$.

In the inference step, only the trained classifier $H$
and feature encoder $G_{\Time\Freq}$ before and after correction are utilized.
When a target samples $\bx^t_i$ to inference is given, \methodname calculates the movement using $d^{ac}_i$ equation followed by a bimodal test and binary classification (known or unknown) is necessary.  An overview of \methodname is in Alg.~\ref{alg:raincoat}; a detailed overview is in Appendix and Alg.~\ref{alg:aac}. 
\section{Experiments}





\subsection{Experimental Setup}
\label{sec:expsetup}

\xhdr{Baselines for Closed-Set DA} 
We consider eight closed-set DA methods. For baselines are general unsupervised DA methods: deep correlation alignment (CORAL)~\cite{dcoral}, CDAN~\cite{cdan}, decision-boundary iterative refinement training with a teacher (DIRT-T)~\cite{Shu2018ADA}, and AdaMatch~\cite{berthelot2022adamatch}. We also consider four unsupervised DA methods for time series: CODATS~\cite{Wilson2020MultiSourceDD}, adversarial spectral kernel matching for unsupervised time series domain adaptation (AdvSKM)~\cite{advskm}, and CLUDA~\cite{Ozyurt2022ContrastiveLF}. We additionally consider source-domain-only training (no transfer) implemented by \cite{adatime}. 
\xhdr{Baselines for Universal DA}
We consider 4 state-of-the-art methods that can reject unknown samples: include UAN~\cite{You2019UniversalDA}, DANCE~\cite{Saito2020UniversalDA}, OVANet~\cite{Saito2021OVANetON}, and UniOT \cite{Chang2022UnifiedOT}. 
\xhdr{Datasets} 
We consider five benchmark datasets from three distinct problem types: (1) human activity recognition: WISDM~\cite{10.1145/1964897.1964918}, HAR~\cite{Anguita2013APD}, HHAR~\cite{Stisen2015SmartDA}; (2) mechanical fault detection: Boiler~\cite{awav-bn36-19}; and (3) EEG prediction: Sleep-EDF~\cite{Goldberger2000PhysioBankPA}. Further details on datasets are given in Appendix~\ref{sec:appendixDataset}. 
\xhdr{Setup for Closed-Set DA}
Individual, participant, or device IDs define domains in the above datasets. Following existing DA research on time series~\cite{Ozyurt2022ContrastiveLF, Wilson2020MultiSourceDD}, we select ten pairs of domains to specify source $\mapsto$ target domains, except for the Boiler dataset where we consider all possible configurations (i.e., six scenarios). 
\xhdr{Setup for Universal DA}
The WISDM dataset is the most challenging because of the considerable label shift across participants. For example, source participant 29 does not perform the activity `jog' at all, but target participant 28 performs `jog' 33\% of the time. To this end, we consider WISDHM to examine the performance of in-dataset UniDA. In addition, HHAR and WISDM contain sensor measurements, and each has one private label (`bike' and `jog'), making them appropriate for cross-dataset evaluation of UniDA.

\xhdr{Evaluation}
We report accuracy and macro-F1 calculated using target test datasets. Accuracy is computed by dividing the number of correctly classified samples by the total number of samples. Macro-F1 is calculated using the unweighted mean of all the per-class F1 scores. It treats all classes equally regardless of their support values. For UniDA, the trade-off between correctly predicting common vs. private classes on the target domain is captured using H-score, defined as the harmonic mean between accuracy on common classes $\textrm{CA}_{c}$ and accuracy on private classes $\textrm{CA}_{u}$, H-score $=(2 \textrm{CA}_{c} \textrm{CA}_{u})/(\textrm{CA}_{c}+ \textrm{CA}_{u})$. The H-score is high only when both $\textrm{CA}_{c}$ and $\textrm{CA}_{u}$ are high. 
{\xhdr{Implementation}
We adopted Adatime's implementation as a benchmarking suite for domain adaptation on time series data \cite{adatime}\footnote{\url{https://github.com/emadeldeen24/AdaTime}}, using 1D-CNN as the encoder because it was suggested to outperform more complex networks such as Resnet and TCN, ensuring differences in performance were attributed to the adaptation algorithm.
}
\begin{figure*}[t]
    \centering
 {\includegraphics[width=0.85\textwidth]{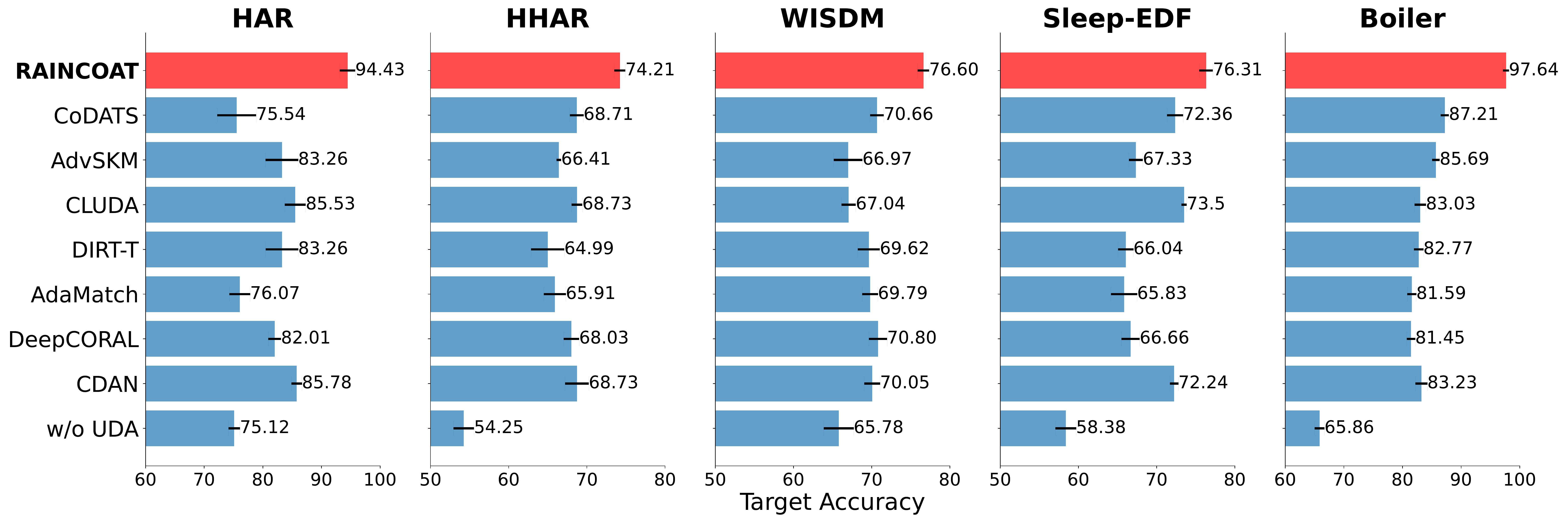}} 
    \caption{Average performance of multiple Closed-set DA methods across multiple datasets. \methodname consistently outperforms all other methods in accuracy on test sets drawn from the target domain dataset. }
    \label{fig:closedDA-Acc}
\end{figure*}
\begin{table}[h!]
\scriptsize
  \caption{H-score of UniDA using WISDM, WISDM$\to$HHAR, HHAR$\to$WISDM, Shown: mean H-score over 5 independent runs. See Table \ref{tab:UniDA-acc} in Appendix for additional results.}
  \label{tab:uni-acc}
  \centering
  \setlength{\tabcolsep}{2pt}
  \begin{tabular}{lcccccc}
    \toprule
    Source $\mapsto$ Target  & UAN & DANCE  & OVANet & UniOT  & \methodname   \\

    \midrule
    WISDM 3 $\mapsto$ 2  & 0 & 0& 0.07& 0.11& \cellcolor{gray!50}\textbf{0.51}\\
    WISDM 3 $\mapsto$ 7 & 0 & 0& 0.2& 0.22& \cellcolor{gray!50}\textbf{0.52}  \\
    WISDM 13 $\mapsto$ 15 & 0& 0.14 & 0.33&0.36 & \cellcolor{gray!50}\textbf{0.50}\\
    WISDM 14 $\mapsto$ 19 & 0.24& 0.28& 0.31& 0.28& \cellcolor{gray!50}\textbf{0.55}\\
    WISDM 27 $\mapsto$ 28 & 0.07& 0.07 & 0.23& 0.35& \cellcolor{gray!50}\textbf{0.59} \\
    WISDM 1 $\mapsto$ 0 & 0.41 & 0.39 & 0.38& 0.40& \cellcolor{gray!50}\textbf{0.43}\\
    WISDM 1 $\mapsto$ 3 & 0.46 &0.49 & 0.45& 0.43& \cellcolor{gray!50}\textbf{0.51}\\
    WISDM 10 $\mapsto$ 11 & 0 & 0 & 0.34& 0.41& \cellcolor{gray!50}\textbf{0.53}\\
    WISDM 22 $\mapsto$ 17 & 0.13 & 0 & 0.32&0.41 & \cellcolor{gray!50}\textbf{0.52}\\
    WISDM 27 $\mapsto$ 15 & 0.43 & 0.51 & 0.46& 0.52& \cellcolor{gray!50}\textbf{0.57}\\

    \hdashline
    WISDM Avg  & 0.17 & 0.19 & 0.31& 0.35& \cellcolor{gray!50}\textbf{0.52}\\
    WISDM Std of Avg & 0.04 & 0.05 & 0.04& 0.05& \cellcolor{gray!50}\textbf{0.04}\\
    \midrule
    W$\to$H 4 $\mapsto$ 0  & 0 & 0.14 & 0.15& 0.19& \cellcolor{gray!50}\textbf{0.49}\\
    W$\to$H 5 $\mapsto$ 1  & 0.24 & 0.22 & 0.25& 0.28& \cellcolor{gray!50}\textbf{0.53}\\
    W$\to$H 6 $\mapsto$ 2  & 0.14 & 0.12 &0.20 &0.25 & \cellcolor{gray!50}\textbf{0.55}\\
    W$\to$H 7 $\mapsto$ 3  & 0 & 0.15 & 0.04& 0.14& \cellcolor{gray!50}\textbf{0.51}\\
    W$\to$H 17 $\mapsto$ 4  & 0.35 & 0.28 &0.41 & 0.45& \cellcolor{gray!50}\textbf{0.57}\\
    W$\to$H 18 $\mapsto$ 5  & 0.20 & 0.27 & 0.29& 0.32& \cellcolor{gray!50}\textbf{0.47}\\
    W$\to$H 19 $\mapsto$ 6  & 0.19 & 0.22 &0.25 &0.28 & \cellcolor{gray!50}\textbf{0.51}\\
    W$\to$H 20 $\mapsto$ 7  & 0.11 & 0.17 &0.35 &0.41 & \cellcolor{gray!50}\textbf{0.49}\\
    W$\to$H 23 $\mapsto$ 8  & 0.21 & 0.28 & 0.47&0.51 & \cellcolor{gray!50}\textbf{0.57}\\
    \hdashline
    W$\to$H Avg & 0.16 & 0.21 & 0.24& 0.28& \cellcolor{gray!50}\textbf{0.52} \\
    W$\to$H Std of Avg & 0.03 & 0.02 & 0.03& 0.02& \cellcolor{gray!50}\textbf{0.02}\\
     \midrule
    H$\to$W 0 $\mapsto$ 4 &  0.23 & 0.28 & 0.33& 0.37& \cellcolor{gray!50}\textbf{0.45}\\
    H$\to$W 1 $\mapsto$ 5 & 0.19 & 0.31 & 0.38&0.42 & \cellcolor{gray!50}\textbf{0.47}\\
    H$\to$W 2 $\mapsto$ 6 & 0.04 & 0.17 & 0.23& 0.29& \cellcolor{gray!50}\textbf{0.39}\\
    H$\to$W 3 $\mapsto$ 7 &  0.25 & 0.32 & 0.34& 0.40& \cellcolor{gray!50}\textbf{0.42}\\
    H$\to$W 4 $\mapsto$ 17  & 0.31 & 0.39 & 0.41&0.40 & \cellcolor{gray!50}\textbf{0.51}\\
    H$\to$W 5 $\mapsto$ 18 & 0.28 & 0.34 & 0.37& 0.36& \cellcolor{gray!50}\textbf{0.48}\\
    H$\to$W 6 $\mapsto$ 19  & 0.42 & 0.42 &0.46 &0.47 & \cellcolor{gray!50}\textbf{0.49}\\
    H$\to$W 7 $\mapsto$ 20  & 0.39 & 0.41 &0.41 & 0.44& \cellcolor{gray!50}\textbf{0.52}\\
    H$\to$W 8 $\mapsto$ 23 & 0.19 & 0.28 & 0.32&0.35 & \cellcolor{gray!50}\textbf{0.46}\\
    \hdashline
    H$\to$W Avg & 0.26 & 0.32 & 0.36&0.39 & \cellcolor{gray!50}\textbf{0.47}\\
    H$\to$W Std of Avg & 0.05 & 0.05 &0.03 & 0.04& \cellcolor{gray!50}\textbf{0.03}\\
    \bottomrule
    \multicolumn{5}{l}{\scriptsize Higher H-score is better. Best performance is indicated in bold.}
  \end{tabular}
  \label{tab:unida-Hscore}
\end{table}

\subsection{Results}
\label{sec:results}
\xhdrd{Q1: How effective is \methodname for closed-set DA?}
Figure \ref{fig:closedDA-Acc} shows each method's average accuracy and standard deviation for selected source-target domain pairs on all datasets. Full results are given in Table \ref{tab:closed-acc} (accuracy) and Table \ref{tab:closed-f1} (Macro-F1). Overall, \methodname has won 5 out of 5 tests (2 metrics in 5 datasets) and makes an average improvement of accuracy (6.77\%) and Macro-F1 (9.00\%) over with the strongest baseline across datasets. Specifically, \methodname improves prediction accuracy by 8.65\% on HAR, 5.48\% on HHAR, 5.8\% on WISDM, 2.81\% on Sleep-EDF, and 10.43\% on Boiler over the strongest baseline on each dataset respectively. In particular, \methodname outperforms CLUDA, the state-of-the-art closed-set DA method for time series, by 8.23\%  (accuracy) and 10.00\% (Macro-F1) averaged over all datasets. \methodname captures and aligns time-frequency features across domains which improve knowledge transfer among time series in the presence of feature shift. 


\xhdrd{Q2: How effective is \methodname for UniDA?}
We report the average H-score in Table \ref{tab:unida-Hscore} and the average accuracy results in Appendix \ref{tab:UniDA-acc}. Results show that \methodname consistently outperforms baselines and achieves state-of-the-art results on DA for time series under both feature and label shift. We note that changes in features and labels of time series data are different from other types of data, such as images, which cause a decrease in the performance of baseline models.
However, \methodname has a significant average improvement over the strongest baseline by 16.33\% (H-score) across datasets with large gaps. This can be attributed to its time-frequency feature encoder and detection of unknown samples via  discriminative features learned using the 'align-and-correct' strategy.


\xhdr{Ablation Studies}
%
%
\begin{table*}[t]
\centering
\caption{Ablation analysis of \methodname. Specifically, the frequency encoder, Sinhorn Alignment, and Correct Step modules are shown below. When no component is checked (first row), it refers to the source-only model. We evaluate \methodname on both closed-set and universal DA and also include average accuracy across all 10 scenarios (source $\mapsto$ target domain) on the WISDM dataset. 
}
\setlength{\tabcolsep}{4.5pt}
\resizebox{1.00\linewidth}{!}{
\begin{tabular}{cccc ccccc|ccccc}
\toprule  
& \multicolumn{3}{c}{Element of \methodname} & \multicolumn{5}{c}{Closed Set DA} & \multicolumn{5}{c}{Universal DA}\\
& Frequency Encoder  &  Sinkhorn &  Correct   & 4 $\mapsto$ 15 & 7 $\mapsto$ 30 & 12 $\mapsto$ 17 & 12 $\mapsto$ 19   & Avg (10 scenarios) &   1 $\mapsto$ 0 & 10 $\mapsto$ 11 & 22 $\mapsto$ 17 & 27 $\mapsto$ 15 & Avg (10 scenarios)\\
\midrule 
1 & ~ & ~ & ~  &  79.86 & 89.32 & 71.53 & 54.29 &65.78  & 64.58 & 54.38 & 42.98 & 38.04 & 40.84 \\
2 & \cmarkb &  & ~ & 89.72 & 90.12 & 84.34 & 83.87 & 75.22 & 70.84 & 65.04 & 44.81 & 54.39 & 42.97\\
3 & ~ & \cmarkb & ~  & 82.43 & 89.88 & 83.14 & 76.74 & 69.66 & 65.13 & 57.44 & 45.14 & 42.42 & 41.25\\
4 & \cmarkb & \cmarkb & & 95.34  & 92.36 & 86.84 & 84.11&  76.24 & 73.68 & 72.37 & 40.79 & 58.17 & 44.08\\

5 & \cmarkb & ~ & \cmarkb  &  90.84 & 90.01 & 86.31 & 79.84 & 76.04 & 74.34 & 66.10 & 48.01 & 57.22 & 46.52\\
6 & \cmarkb & \cmarkb & \cmarkb & 97.91 & 91.28 & 89.80 & 85.00 & 76.60 &  82.57 & 76.36 & 48.16 & 66.42 & 53.51\\
\bottomrule 
\end{tabular}
}

\label{tab:ablation}
\end{table*}
Next, we present the setup and results of our ablation studies discussed in Section \ref{sec:results}. We study the following questions \textbf{Q1:} How effective is the time-frequency encoder? \textbf{Q2:} Will the correct step decrease the performance when there is no label shift? \textbf{Q3:} Is Sinkhorn divergence a better measurement for our time-frequency feature?  We evaluate how relevant the model components are for effective DA. We perform the ablation study using WISDM since it is a more challenging dataset and present results in Table \ref{tab:ablation}. 
When no component is used (\nth{1} row in Table \ref{tab:ablation}), it refers to a source-only model. When Sinkhorn is not used (\nth{2},\nth{5} row in Table \ref{tab:ablation}), we use MMD to align features. It can be observed that using the frequency encoder alone (\nth{2} row) results in performance improvement (accuracy) of 9.44\% for Closed-set DA and 2.33\% for UniDA on average. It demonstrates the effectiveness of a frequency encoder for handling the feature shift of time series. When the frequency encoder (\nth{2} row) is further equipped with a correction step (\nth{5} row), it verifies the effectiveness of the correction step when there is a label shift. By comparing the \nth{5} row with the \nth{2} and \nth{4} row, we find that the correction step does not lead to a performance drop for Closed-set DA. This finding indicates that \methodname is suitable for resolving both feature and label shifts, even if no prior information on feature and label shifts is given. By comparing \nth{2} row with \nth{4} row, we observe Sinkhorn Divergence brings consistent improvement for both Closed-set DA (1.02\%) and UniDA (1.11\%), which demonstrates the benefit of Sinkhorn Divergence for aligning frequency features. 

We systematically investigate the role of Sinkhorn divergence in \methodname to align time-frequency features. This analysis is particularly relevant because existing methods do not consider frequency features as a potential source of feature shifts. To numerically verify that Sinkhorn divergence is an appropriate divergence measurement for aligning time-frequency features, we conducted experiments by replacing the encoder with our time-frequency feature encoder for additional baselines. We select four  representative and strong baselines to ensure a diverse category of adaptation methods: CoDATS, DeepCoral, AdvSKM, and CLUDA. We run Closed-Set DA experiments on HAR datasets and report the average prediction accuracy in Table \ref{tab:encoder}. The results demonstrate that the time-frequency feature encoder achieves the highest accuracy when combined with Sinkhorn divergence, highlighting the effectiveness of using this method to align time-frequency features for time series. Furthermore, all methods show improved prediction accuracy when using our time-frequency encoder, indicating that leveraging and aligning both time and frequency features are crucial for domain adaptation in time series. Additional experiments on \methodname's loss function and sample complexity are in Table \ref{tab:weight} and \ref{tab:sample} in Appendix \ref{sec:ablation}.


\begin{table}[ht]
\small
\centering
\caption{Accuracy comparison of different domain adaptation methods with and without a time-frequency encoder on the HAR dataset. Results on `\methodname without our encoder` indicate performance when only Sinkhorn divergence is used.}
\label{tab:encoder}
\begin{tabular}{ c|c|c }
\toprule
Method & W/o our encoder & W/ our encoder  \\
\midrule
CoDATS & 75.54 & \textbf{83.67} \\
DeepCoral & 82.01 & \textbf{89.75}\\
AdvSKM & 83.26 & \textbf{89.64}\\
CLUDA  & 85.53 & \textbf{90.62}\\
\methodname & 82.48 & \textbf{\textit{94.43}} \\
\bottomrule
\end{tabular}
\end{table}

\section{Conclusion}
We introduce \methodname, a domain adaptation approach for time series that addresses both feature and label shifts. \methodname combines time and frequency space features, aligns them across domains, corrects misalignments, and detects label shifts. Experimental results on five datasets demonstrate \methodname's effectiveness, achieving up to 6.7\% improvement on closed-set domain adaptation and 16.33\% improvement on universal domain adaptation. 


\section*{Acknowledgements}

We gratefully acknowledge the support of the Under Secretary of Defense for Research and Engineering under Air Force Contract No.~FA8702-15-D-0001 and awards from NIH under No.~R01HD108794, Harvard Data Science Initiative, Amazon Faculty Research, Google Research Scholar Program, Bayer Early Excellence in Science, AstraZeneca Research, and Roche Alliance with Distinguished Scientists. Any opinions, findings, conclusions or recommendations expressed in this material are those of the authors and do not necessarily reflect the views of the funders. The authors declare that there are no conflict of interests.

\bibliographystyle{unsrtnat}
\bibliography{main}  
\newpage
\appendix
\onecolumn

\section{Further Information on Domain Alignment of Time-Frequency Feature}
\label{sec:appendixMMD}

We first show the distributions of Fourier amplitude and phase. 
\begin{equation}
\begin{aligned}
f(a, p) &=a \times f(x=a \sin p, y=a \cos p) \\
&=a \times \frac{1}{2 \pi} \cdot \exp \left(-\frac{a^2\left(\sin ^2 \theta+\cos ^2 \theta\right)}{2}\right) \cdot \\
&=\frac{a}{2 \pi} \cdot \exp \left(-\frac{a^2}{2}\right) \cdot 
\mathbb{I} \\
&=a \cdot \exp \left(-\frac{a^2}{2}\right) \cdot \mathbb{I}(a \geqslant 0) \times \frac{1}{2 \pi} \cdot \\
&=\text { Rayleigh }(a \mid 1) \cdot \mathrm{U}(p \mid 0,2 \pi) \\
&\quad (a \geqslant 0,0 \leqslant p \leqslant 2 \pi).
\end{aligned}
\end{equation}
We can observe the amplitude can be arbitrarily large, and thus $\ba^s$ and $\ba^t$ might have a disjoint set when the frequency feature shift is significant. As a result, we need to consider a measurement that can measure the distance of two arbitrary distributions.

\xhdr{Sinkhorn Divergence} 
We consider two discrete probability measures represented as sums of weighted Dirac atoms:
\begin{equation}
{\bmu}=\sum_{i=1}^n {\bmu}_i \boldsymbol{\delta}_{{\bz}_i} \text { and } {\bnu}=\sum_{j=1}^m {\bnu}_j \boldsymbol{\delta}_{{\bz}_j}
\end{equation}
Here, $\bmu \in R_{+}^n$ and $\bnu \in R_{+}^m$ are non-negative
vectors of length $n$ and $m$ that sum up to 1. We denote their probabilistic couplings, set $\boldsymbol{\Pi}$ and cost matrix $\bC$, as:
\begin{equation}
\begin{array}{cc}
     &\boldsymbol{\Pi}({\bmu}, {\bnu})=\left\{\mathbf{P} \in \mathbb{R}_{+}^{n \times m}, \mathbf{P} \mathbf{1}_m={\bmu}, \mathbf{P}^{\top} \mathbf{1}_n={\bnu}\right\}
\\
 &\bC=(\bC_{i j}) \in \mathbb{R}_{+}^{n \times m}, \left.{\bC}_{i j}=\left\|{\bz}_i-{\bz}_j\right\|^p\right.
\end{array}
\end{equation}
Sinkhorn divergence \cite{cuturi2013sinkhorn,cuturi2016smoothed} was proposed as an entropic regularization of the  Wasserstein distance \cite{Fournier2013OnTR} that interpolates between the pure OT loss for $\eta=0$ and MMD \cite{JMLR:v13:gretton12a} losses for $\eta\to\infty$ and offers a computationally efficient way to approximate OT costs. It thus provides a good tradeoff between (a) favorable sample complexity and unbiased gradient estimates and (b) non-flat geometry of OT \citep{Genevay:2018, Feydy:2019}.  The Sinkhorn divergence between $\bmu$ and $\bnu$ is given by
\begin{equation}
\mathcal{S}_\eta(\bmu, \bnu)=\min _{\mathbf{P} \in \Pi(\bmu, \bnu)}\{\langle\boldsymbol{C}, \mathbf{P}\rangle + \eta H(\mathbf{P})\},
\end{equation}
where $H(P) =  \sum_{i,j} \mathbf{P}_{ij}\text{log}(\mathbf{P}_{ij})$ is the negative entropy and  $\eta>0$ is a regularization parameter.  By making 
$\eta$ higher, the resulting coupling matrix will be smoother, and as $\eta$ goes to zero, it will be sparser, with the solution being close to the optimal transport solution. The Sinkhorn algorithm to find such a coupling matrix is efficiently provided in Alg. \ref{alg:sinkhorn}.

\begin{figure*}
    \centering
    \subfigure[]{\includegraphics[width=0.24\textwidth]{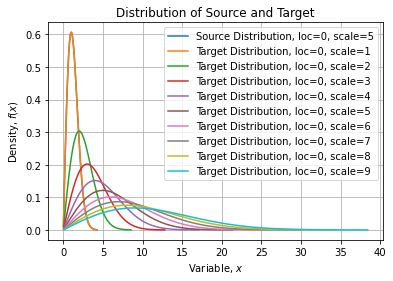}} 
    \subfigure[]{\includegraphics[width=0.24\textwidth]{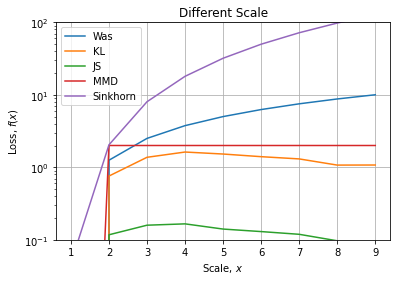}} 
    \subfigure[]{\includegraphics[width=0.24\textwidth]{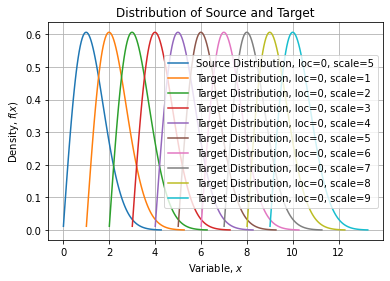}}
    \subfigure[]{\includegraphics[width=0.24\textwidth]{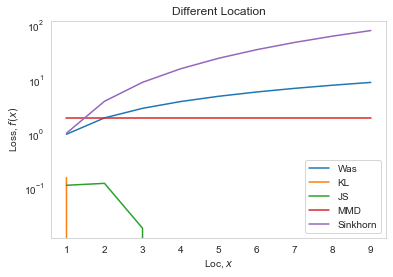}}
    \caption{ (a) Rayleigh distributions with different scales (b) JSD and KLD  (c) Rayleigh distributions with different locations (d) JSD and KLD. The Mean Squared Error (MSE) measure exhibits a rapid increase and explodes when there is a significant location shift between distributions. As shown in Figure b, using the Kullback-Leibler (KL) divergence helps mitigate this issue, but it is still not bounded. Furthermore, as depicted in Figure c, both Jensen-Shannon Divergence (JSD) and KL divergence struggle to provide meaningful gradients when there is a substantial location shift. This observation is consistent with the fact that JSD cannot offer usable gradients when distributions are supported on non-overlapping domains, as explained in \cite{kolouri2018sliced}. The Wasserstein distance demonstrates a linear relationship with the shift, but it also lacks a bound. 
    }
    \label{fig:sinkhorn_motivation}
\end{figure*}

 Optimal transport losses have appealing geometric properties, but it takes $O(n^3\log n)$ to compute. On the other hand, discrepancy metrics such as MMD are geometry-aware and can scale up to large batches with a low sample complexity. But we realize that measuring the discrepancy of frequency features using Sinkhorn has a stronger Gradient than MMD. Specifically, consider MMD with an RBF kernel, the gradient of MMD w.r.t. a particular sample $\mathbf{\bz^s}$
is $\nabla_{\bz^s} D_{M M D}(\mathbf{Z}^s, \mathbf{Z}^t)=\frac{1}{N^2} \sum_j k\left(\bz_i^s, \bz_j^s\right) \frac{\bz_j^s-\bz_i^s}{\sigma^2}-\frac{2}{N M} \sum_j k\left(\bz_i^s, \bz_j^t\right) \frac{\bz_j^t-\bz_i^s}{\sigma^2}$. When minimizing MMD,
the first term is a repulsive term between the samples from $p(\bz^s)$, and the second term is an attractive
term between the samples from $p(\bz^s)$ and $p(\bz^t)$. The L2 norm of the term between two samples $\bz^s$ and $\bz^t$ is small if  $\left\lVert \bz^s-\bz^t\right\rVert_2$ is either too small or too large. This is saying if $p(\bz^s)$ is far away from $p(\bz^t)$, the model will not receive strong gradients (bounded by a  small norm). From another viewpoint, \cite{Feydy:2019} demonstrated that the norm of MMD strongly relies on the smoothness of the reference measure and tends to have vanishing gradients when points of the measures' support are disjoint. 
Now let's look at the gradients of Sinkhorn. Denote a Lipschitz cost function as $\mathrm{C}(\bz^s, \bz^t)$. For $\eta>0$, the associated Gibbs kernel is defined through
$$
k_{\eta}:(\bz^s, \bz^t) \in \mathcal{Z}^s \times \mathcal{Z}^t \mapsto \exp (-\mathrm{C}(\bz^s, \bz^t) / \eta)
$$
\cite{Feydy:2019} show that the Sinkhorn divergence gradient w.r.t a particular sample $\mathbf{\bz^s_i}$ is largely determined by the magnitude of:
\begin{equation}
\eta \left(\log(\exp(-C(\bz_i^s,\bz^s_j))/\eta))-\log(\exp(-C(\bz_i^s,\bz^t_j))/\eta))\right).
\end{equation}
Different from MMD, the cost function $C(\bz^s, \bz^t)$ replaces the Euclidean distance with an absolute distance $|\bz^s_i-\bz^t_j|$. Then, the gradient is always strong regardless of the closeness between $\bz^s_i$ and $\bz^t_j$.
To numerically verify this claim, we compare the magnitude of the
gradients of different shifts in Figure \ref{fig:sinkhorn_motivation}. It shows that Sinkhorn has stronger gradients than alternative approaches. 

\begin{algorithm}

  \caption{Simplified illustration of computation of Sinkhorn Divergence \cite{Sinkhorn1964ARB} 
    }
    \label{alg:sinkhorn} 
  \begin{algorithmic}[1]
    \Function{Sinkhorn Divergence}{$z^s, z^t$} 
      \Let{$a, b$}{$\mathbf{1}_n/n, \mathbf{1}_n/n$}
      \Let {$\boldsymbol{C}$}{$\|z^s-z^t\|^p$}
      \Let{$\boldsymbol{K}$}{$\text{exp}^{(-C/\eta)}$}
        \For{$j \gets 1 \textrm{ to } J$}
        \State $\boldsymbol{a}^{(j)} \leftarrow \boldsymbol{\mu} \oslash \boldsymbol{K} \boldsymbol{b}^{(j-1)} ; \boldsymbol{b}^{(j)} \leftarrow \boldsymbol{\nu} \oslash \boldsymbol{K}^{\top} \boldsymbol{a}^{(j-1)}$
        \EndFor
        \Let{$\mathcal{L}_{align}$}{$\sum \boldsymbol{C}\textrm{diag}(\boldsymbol{a}^{(j)})\boldsymbol{K} \textrm{diag}(\boldsymbol{b}^{(j-1)})$}
      \State 
      \Return{$\mathcal{L}_{align}$}
    \EndFunction
  \end{algorithmic}
\end{algorithm}




\section{Details on Neural Networks \methodname Algorithm}
\xhdr{Encoder}
\label{asec:encoder}
An aspect that has not been adequately emphasized is the composition of a practical time series, which consists of a blend of numerous oscillations at various frequencies, potentially even infinite. In order to address this, \methodname takes into account both time and frequency characteristics during encoding. We use DFT in our work and leave other approaches for future work. To mitigate the issue of frequency leakage, \methodname incorporates a smoothing process on the input $\bx_i$. The selection of a suitable smoothing function presents several options, yet the distinctions between them are often negligible in practical applications. In our approach, we utilize either the cosine or Hann window $w$ as a smoothing function, commonly known as tapering functions. These functions are designed using a raised cosine with optimized non-zero endpoints to minimize the impact of nearby side lobes.  It is defined as: 
\begin{equation}
\label{eqn:hann}
w[n]=0.5-0.5 \cos \left(\frac{2 \pi n}{N-1}\right) \quad 0 \leq n \leq N-1
\end{equation}

Following the smoothing process, the smoothed signal $\bx_i$ undergoes the Discrete Fourier Transform (DFT), resulting in the transformed vector $\bv_i$. However, the direct implementation of DFT, as shown in Equation \eqref{eqn:DFT}, can be computationally inefficient for long signals. To address this, we can leverage the fast Fourier transform (FFT) algorithm \cite{Cooley1965AnAF} to scale up the computation efficiently. An important property of the Fourier domain representation of real signals is the Hermitian property: $\bv[m] = \bv[-m]$. This property implies that we can save memory by storing only the one-sided representation containing positive frequencies. By doing so, we can reduce memory requirements by half. For a comprehensive understanding of DFT, please refer to \citet{10.5555/557358}.

Next, \methodname applies a convolution operator specifically on the \textbf{low-frequency modes} of $\bv_i$, which aligns with existing approaches in neural network-based frequency analysis. The rationale behind this step is that by focusing on low-frequency modes, the operator smooths out high-frequency details that often exhibit less structure compared to their low-frequency counterparts. This process helps to preserve the low-rank structure of signals, facilitating alignment. Unlike previous works such as \cite{li2021fourier, fedformer}, \methodname does not incorporate an additional linear transform, as this step is employed to preserve time-space features. Instead, \methodname adopts a time feature encoder.

Subsequently, we extract the amplitude and phase from the output of the convolutional layer, as we have observed that representing these features in polar coordinates (amplitude-phase representation) tends to be more domain-invariant and introduces a useful inductive bias into the model.

Finally, the extracted frequency features are concatenated with the time features. Various approaches, such as manifold alignment and self-attention, can be explored for feature fusion in future work. However, we defer these investigations to our future research.

The process of frequency-space feature extraction is as follows: given a time series $\bx$, it is first multiplied by the Hann window function \eqref{eqn:hann} to mitigate frequency leakage. Subsequently, a convolution operation is applied to the smoothed signal. This results in the frequency-space feature $e_{\mathcal{F}}$, which is obtained by concatenating the results using \eqref{eqn:ap}. For the extraction of time-space features, any appropriate network architecture can be utilized. In this work, we adopt a Convolutional Neural Network (CNN) to ensure a fair comparison with existing studies. The pseudocode for the time-frequency feature extraction is presented in Algorithm \ref{alg:freq-align}.

\xhdr{Fourier Neural Operator}
 Fourier neural operator (FNO)~\cite{li2021fourier} performs temporal predictions by combining the Fourier transform with neural networks. Define a convolution operator ``$\ast$" and weight matrix $\mathbf{B}$ , the Fourier layer in FNO can be summarized as:
$$
\begin{array}{ll}
\text { (1) DFT} & \bv =[\operatorname{DFT}(\bx)] \\
\text { (2) Frequency Convolution } & \tilde{\be}_{\mathcal{F}}= \mathbf{B} \ast \bv \\
\text { (3) IDFT} & \tilde{\bx}=[\operatorname{IDFT}(\tilde{\be}_{\mathcal{F}})]
\end{array}
$$
 FNO then adds the output of the Fourier layer with the bias term (a linear transformation) and applies the activation function. \methodname differs a lot from FNO. The only shared component is the frequency convolution, as we mentioned previously. 
 
\xhdr{Decoder}
In order to acquire discriminative features, \methodname employs a decoder that is trained through a reconstruction task. Given a latent representation $\bz_i$ obtained from either the source or target, we decompose it into frequency and time features. By performing separate reconstructions on both $\be_{\Freq}$ and $\be_{\Time}$, we can easily reconstruct the original signal $\bx$. For the frequency feature reconstruction, we apply the inverse Discrete Fourier Transform (DFT) on outputs of Convolutional Fourier Transform, while for the time feature reconstruction, a standard deconvolution network is utilized on $\be_{\Time}$. These reconstructed frequency and time components are then combined to form $\hat{\bx}$, an approximation of the original signal. To train the reconstruction task, we employ the L1 loss function.

\xhdr{Prototypical Classifier}
In \methodname, we employ a prototypical classifier, inspired by the work of \citet{Kim2019VariationalPO}. The normalized feature vector $\bz$ serves as the input to the classifier $H$. The classifier $H$ consists of weight vectors $\bW = [\bw_1, \bw_2, \cdots, \bw_{C}]$, where $C$ denotes the number of classes. These weight vectors can be interpreted as estimated prototypes for each class. By utilizing the prototypical classifier, \methodname aims to classify input samples based on their similarity to the estimated prototypes. The feature vector $\bz$ is compared to each weight vector $\bw_i$, and the class with the closest prototype is assigned as the predicted class label. This approach leverages the discriminative power of the prototypes to facilitate accurate classification.

\begin{algorithm}
 
  \caption{Time-Frequency Feature Encoder and Decoder, Domain Alignment via Sinkhorn Divergence 
   }
   \label{alg:freq-align} 
  \begin{algorithmic}[1]
    \Function{Time-Freq Encoder ${G}_{\Time\Freq}$}{$x$} 
    \Let{$x$}{\text{smooth}($x$), as shown in \eqref{eqn:hann}}
      \Let{$\bv_{\Freq}$}{$DFT(x)$}
      \Let{$\title{\bx_{\Freq}}$}{\textproc{Spec-conv}($\bv_{\Freq}$)} 
      \Let {$\ba$, $\bp$}{$\bv_{\Freq}|, atan2(Im(x_{\Freq}), Re(x_{\Freq}))$}
    \Let{$\be_{\Freq}$}{$\textproc{Concat}(\ba, \bp)$}
      \Let{$\be_{\Freq}$}{\textproc{Time-conv}($x$)} 
    \Let{$\bz$}{$\textproc{Concat}(\be_{\Freq}, \be_{\Freq})$}
      \State \Return{$\bz$}
    \EndFunction
\State
    \Function{Time-Freq Decoder $\mathcal{U}_{\mathcal{T}\mathcal{F}}$}{$z$} 
      \Let {$e_{\mathcal{T}}, e_{\mathcal{F}}$}{$z$}
        \Let{$\bar{x}_{\mathcal{T}}, \bar{x}_{\mathcal{F}}$}  {$\textproc{ConvTran1d}(e_{\mathcal{T}}), \textproc{IFFT}(e_{\mathcal{F}}) $}
        \Let {$\bar{x}$}{$\bar{x}_{\mathcal{T}}+\bar{x}_{\mathcal{F}}, $}
      \State \Return{$\bar{x}$}
    \EndFunction

  \end{algorithmic}
\end{algorithm}

\begin{algorithm}
\caption{Detailed overview of \methodname }
\label{alg:aac}
\begin{algorithmic} 
\State \textbf{Input}: dataset $\mathcal{D}_s$ , $\mathcal{D}_t$ ; epochs $E_1, E_2$
\State \textbf{Initialization}: Parameter $\Gamma$ for Time-Frequency Feature Encoder $G_{\mathcal{T}\mathcal{F}}$, $\Phi$ for Time-Frequency Feature Decoder $U_{\mathcal{T}\mathcal{F}}$, weight vectors $W = [w_1, w_2, \cdots, w_{C^s}]$ for prototypical classifier,.
\State \textit{Stage 1: Align, introduced in Section \ref{sec:tf-encoder} \ref{sec:align}}
       \For{$e \gets 1 \textrm{ to } E_1$}
        \While{$\mathcal{D}_t$ not exhausted}
            \State Sample $x^s, y^s$ from $\mathcal{D}_s$, $x^t$ from $\mathcal{D}_t$
            \State Extract: $z^s  \gets G_{\mathcal{T}\mathcal{F}}(x^s)$ (use Algorithm \ref{alg:freq-align})
            \State Extract: $z^t  \gets G_{\mathcal{T}\mathcal{F}}(x^t)$ (use Algorithm \ref{alg:freq-align})
            \State Reconstruct $\bar{x}^{s} \gets \mathcal{U}_{\mathcal{T}\mathcal{F}}(z^s)$
            \State Compute:$\mathcal{L}_{align} = \textproc{Sinkhorn}(z^s, z^t, \epsilon)$
            \State \Comment{in algorithm \ref{alg:freq-align}}
            \State Compute:$\mathcal{L}_{recon} = |x^s-\bar{x}^{s}|$
            
            \State Predict: $\hat{y_s} = \textproc{Classifier}(z^s)$
            \State Compute $\mathcal{L}_{cls}= CE(y^s, \hat{y^s})$
            \State $\mathcal{L}_{total}=\mathcal{L}_{recon}+\mathcal{L}_{align}+\mathcal{L}_{cls}$
            \State Update $\Gamma,\Phi,W$ with $\nabla\mathcal{L}_{total}$
        \EndWhile
        \EndFor
\State
\State \textit{Stage 2: Correct}, introduced in Sec. \ref{sec: correct}
\State Compute distance to prototypes before correct: 
\State \quad $d_{align}=\frac{\mathbf{Z}^s \cdot W}{\|\mathbf{Z}^s\|\|\mathbf{W}\|}$
\For{$e \gets 1 \textrm{ to } E_2$}
        \While{$\mathcal{D}_t$ not exhausted}
            \State Sample $x^t$ from $\mathcal{D}_t$
            \State Extract: $z^t  \gets G_{\mathcal{T}\mathcal{F}}(x^t)$ (use Algorithm \ref{alg:freq-align})
            \State Reconstruct $\bar{x}^{t} \gets \mathcal{U}_{\mathcal{T}\mathcal{F}}(z^t)$
            \State Compute:$\mathcal{L}_{recon} = |x^s-\bar{x}^{s}|$
            \State Update $\Gamma,\Phi$ with $\nabla\mathcal{L}_{recon}$
        \EndWhile
        \EndFor
\State Compute distance to prototypes after correct: 
\State \quad $d(\mathbf{Z},\mathbf{W})=\frac{\mathbf{Z}^s \cdot W}{\|\mathbf{Z}^s\|\|\mathbf{W}\|}$
\State
\State \textit{Stage 3: Inference}, introduced in Sec. \ref{sec:inference}
\State Compute drift during correct: 
\State \quad $drift=|d_{correct}-d_{align}|$
\For {$c \gets 1 \textrm{ to } C$}
    \State Compute DIP statistic: $dip=\textproc{DIP}(\{drift\}_{y=c})$
    \If{$dip<0.05$}
    \Comment{Two modes detected}
        \State $\mu_c^{common}, \mu_c^{private} =\textproc{K-Means}(\{drift\}_{y=c})$
    \EndIf

\EndFor
\end{algorithmic}
\end{algorithm}


\section{Additional Experimental Results}

\subsection{Dataset Details}
\label{sec:appendixDataset}
We evaluate the performance of \methodname on five benchmark datasets, each with its own characteristics. The datasets we consider are as follows:

\textbf{(1) WISDM} \cite{10.1145/1964897.1964918}: This dataset consists of 3-axis accelerometer measurements obtained from 30 participants. The measurements are collected at a frequency of 20 Hz. To predict the activity (label) of each participant during specific time segments, we utilize non-overlapping segments of 128-time steps. The dataset includes six activity labels: walking, jogging, sitting, standing, walking upstairs, and walking downstairs.

\textbf{(2) Boiler}~\cite{awav-bn36-19}: The dataset comprises sensor data from three boilers recorded between March 24, 2014, and November 30, 2016. Each boiler is treated as a separate domain. The objective of this task is to detect mechanical faults specifically related to the blowdown valve of each boiler.

\textbf{(3) HAR}~\cite{Anguita2013APD}: This dataset contains measurements from a 3-axis accelerometer, 3-axis gyroscope, and 3-axis body acceleration. The data is collected from 30 participants at a sampling rate of 50 Hz. Similar to the WISDM dataset, we use non-overlapping segments of 128-time steps for classification. The goal is to classify the time series into six activities: walking, walking upstairs, walking downstairs, sitting, standing, and lying down.

\textbf{(4) HHAR}~\cite{Stisen2015SmartDA}: This dataset consists of 3-axis accelerometer measurements from 30 participants. The measurements are captured at a frequency of 50 Hz. Non-overlapping segments of 128-time steps are used for classification purposes. The dataset includes six activity labels: biking, sitting, standing, walking, walking upstairs, and walking downstairs.

\textbf{(5) Sleep-EDF}~\cite{Goldberger2000PhysioBankPA}: This dataset contains electroencephalography (EEG) readings from 20 healthy individuals. The objective is to classify the EEG readings into five sleep stages: wake (W), non-rapid eye movement stages (N1, N2, N3), and rapid eye movement (REM). In line with prior research, we focus on the Fpz-Cz channel for our analysis.

For detailed statistics regarding each dataset, please refer to Table \ref{tab:dataset}. These datasets cover a range of applications and challenges, allowing us to evaluate the effectiveness and robustness of \methodname across various domains.

\begin{table}[h]
  \caption{Summary of datasets.}
  \label{tab:dataset}
  \centering
  \small
 {
  \begin{tabular}[c]{l|cccccc}
  \toprule
    Dataset & \#Subjects & \#Channels & Length &\# Class & \#Train & \# Test\\
    \midrule
    HAR   & 30 & 9 & 128 & 6 & 2,300 & 990\\
    HHAR  & 9  & 3 & 128 & 6 & 12,716 & 5,218\\
    WISDM & 30 & 3 & 128 & 6 & 1,350 & 720 \\
    Sleep-EDF  & 20  & 1 & 3,000 & 5 & 14,280 & 6,310\\
    Boiler  &  3 & 20 & 36 & 2 & 160,719 & 107,400\\
    \bottomrule
  \end{tabular}
  }
\end{table}

\subsection{Experimental Details}
\label{ssec: expdetail}
In this section, we provide implementation details of \methodname and the baseline methods. The implementation was done in PyTorch, based on the code available at \href{https://github.com/emadeldeen24/AdaTime}{here}. The experiments were conducted on a NVIDIA GeForce RTX 3090 graphics card.

To ensure fair comparisons, we carefully selected the appropriate encoder and scale across all methods. This consideration was applied to all our comparisons. For the extraction of time-space features, we utilized a 1D-convolutional neural network (CNN) as the encoder. This configuration was kept consistent across all methods to ensure a fair comparison, where differences in prediction performance could be attributed to the adaptation algorithm itself. The implementation of the 1D-CNN architecture was adapted from a recently published benchmark codebase in the literature \cite{adatime}, which has also been employed by others \cite{Ozyurt2022ContrastiveLF}. The 1D-CNN architecture consists of three blocks, each consisting of a 1D convolutional layer, followed by a 1D batch normalization layer, a rectified linear unit (ReLU) function for non-linearity, and finally, a 1D max-pooling layer. Extensive benchmark evaluations have demonstrated that the 1D-CNN consistently outperforms more complex backbone networks, such as 1D-Resnet-18 and TCN, hence our choice of the 1D-CNN encoder.

During model training, we employed the Adam optimizer for all methods, with carefully tuned learning rates specific to each method. The hyperparameters of Adam were selected after conducting a grid search on source validation datasets, exploring a range of learning rates from $1 \times 10^{-4}$ to $1 \times 10^{-1}$. The learning rates were chosen to optimize the performance of each method.

Key hyperparameters for \methodname are reported in Tables \ref{tab:exp-har}, \ref{tab:exp-eeg}, \ref{tab:exp-wisdm}, \ref{tab:exp-hhar}, and \ref{tab:exp-boiler}. The Fourier Frequency modes used for HAR, EEG, HHAR, WISDM, and Boiler datasets are 64, 200, 64, 64, and 10, respectively. For the regularization term used in the Sinkhorn divergence, we consistently used a value of $1 \times 10^{-3}$ across all datasets and experiments. Additional hyperparameters can be found in the codes.

By providing these implementation details and hyperparameter values, we ensure transparency and reproducibility of the experiments conducted with \methodname.

\begin{table}
    \centering
    \begin{tabular}{c|c|c|c}\toprule
          Method & Epoch & Batch Size & Learning rate \\\midrule
          CoDATS  &   50 & 32 &  $1e-3$ \\
          AdvSKM  &   50 & 32 &  $5e-1$ \\
          CLUDA  &   50 & 32 &  $1e-2$ \\
          DIRT-T  &   50 & 32 &  $5e-4$ \\
          AdaMatch  &   50 & 32 &  $3e-3$ \\
          DeepCoral  &   50 & 32 &  $5e-3$ \\
          CDAN  &   50 & 32 &  $1e-2$ \\
          \methodname  &   50 & 32 &  $5e-4$ \\
        \bottomrule
    \end{tabular}
    \caption{Experimental details for HAR dataset.}
    \label{tab:exp-har}
\end{table}

\begin{table}
    \centering
    \begin{tabular}{c|c|c|c}
        \toprule
        Hyperparameter & Epoch & Batch Size & Learning rate \\
     \midrule
          CoDATS  &   50 & 128 &  $1e-2$ \\
          AdvSKM  &   50 & 128 &  $5e-4$ \\
          CLUDA  &   50 & 128 &  $5e-4$ \\
          DIRT-T  &   50 & 128 &  $5e-4$ \\
          AdaMatch  &   50 & 128 &  $5e-4$ \\
          DeepCoral  &   50 & 128 &  $5e-4$ \\
          CDAN  &   50 & 128 &  $1e-3$ \\
          \methodname  &   50 & 128 &  $1e-3$ \\
          \bottomrule
    \end{tabular}
    \caption{Experimental details for EEG dataset.}
    \label{tab:exp-eeg}
\end{table}

\begin{table}
    \centering
    \small
    \begin{tabular}{c|c|c|c}
        \toprule
        Hyperparameter & Epoch & Batch Size &  Learning rate\\
        \midrule
          CoDATS  &   50 & 64 &  $1e-3$ \\
          AdvSKM  &   50 & 64 &  $3e-4$ \\
          CLUDA  &   50 & 64 &  $1e-3$ \\
          DIRT-T  &   50 & 64 &  $1e-3$ \\
          AdaMatch  &   50 & 64 &  $2e-3$ \\
          DeepCoral  &   50 & 64 &  $5e-2$ \\
          CDAN  &   50 & 64 &  $1e-3$ \\
          \methodname  &   50 & 64 &  $1e-3$ \\
        \bottomrule
    \end{tabular}
    \caption{Experimental details for WISDM dataset}
    \label{tab:exp-wisdm}
\end{table}

\begin{table}
    \centering
    \small
    \begin{tabular}{c|c|c|c}
        \toprule
        Hyperparameter & Epoch & Batch Size &  Learning rate\\
         \midrule
          CoDATS  &   50 & 32 &  $1e-3$ \\
          AdvSKM  &   50 & 32 &  $3e-4$ \\
          CLUDA  &   50 & 32 &  $1e-3$ \\
          DIRT-T  &   50 & 32 &  $1e-3$ \\
          AdaMatch  &   50 & 32 &  $3e-3$ \\
          DeepCoral  &   50 & 32 &  $5e-4$ \\
          CDAN  &   50 & 32 &  $1e-3$ \\
          \methodname  &   50 & 32 &  $1e-3$ \\
        \bottomrule
    \end{tabular}
    \caption{Experimental details for HHAR dataset.}
    \label{tab:exp-hhar}
\end{table}

\begin{table}
    \centering
    \small
    \begin{tabular}{c|c|c|c}
        \toprule
        Hyperparameter & Epoch & Batch Size &  Learning rate \\
         \midrule
          CoDATS  &   30 & 32 &  $5e-4$ \\
          AdvSKM  &   30 & 32 &  $1e-3$ \\
          CLUDA  &   30 & 32 &  $1e-3$ \\
          DIRT-T  &   30 & 32 &  $1e-3$ \\
          AdaMatch  &   30 & 32 &  $3e-3$ \\
          DeepCoral  &   30 & 32 &  $5e-4$ \\
          CDAN  &   30 & 32 &  $1e-3$ \\
          \methodname  &   50 & 32 &  $1e-3$ \\
          \bottomrule
    \end{tabular}
    \caption{Experimental details for Boiler dataset.}
    \label{tab:exp-boiler}
\end{table}

\subsection{
t-SNE Visualizations of Learned Representations for Closed-Set DA}
\begin{figure}[htbp]
  \centering
  \subfigure[HAR 2- HAR 11]{
    \includegraphics[width=0.21\textwidth]{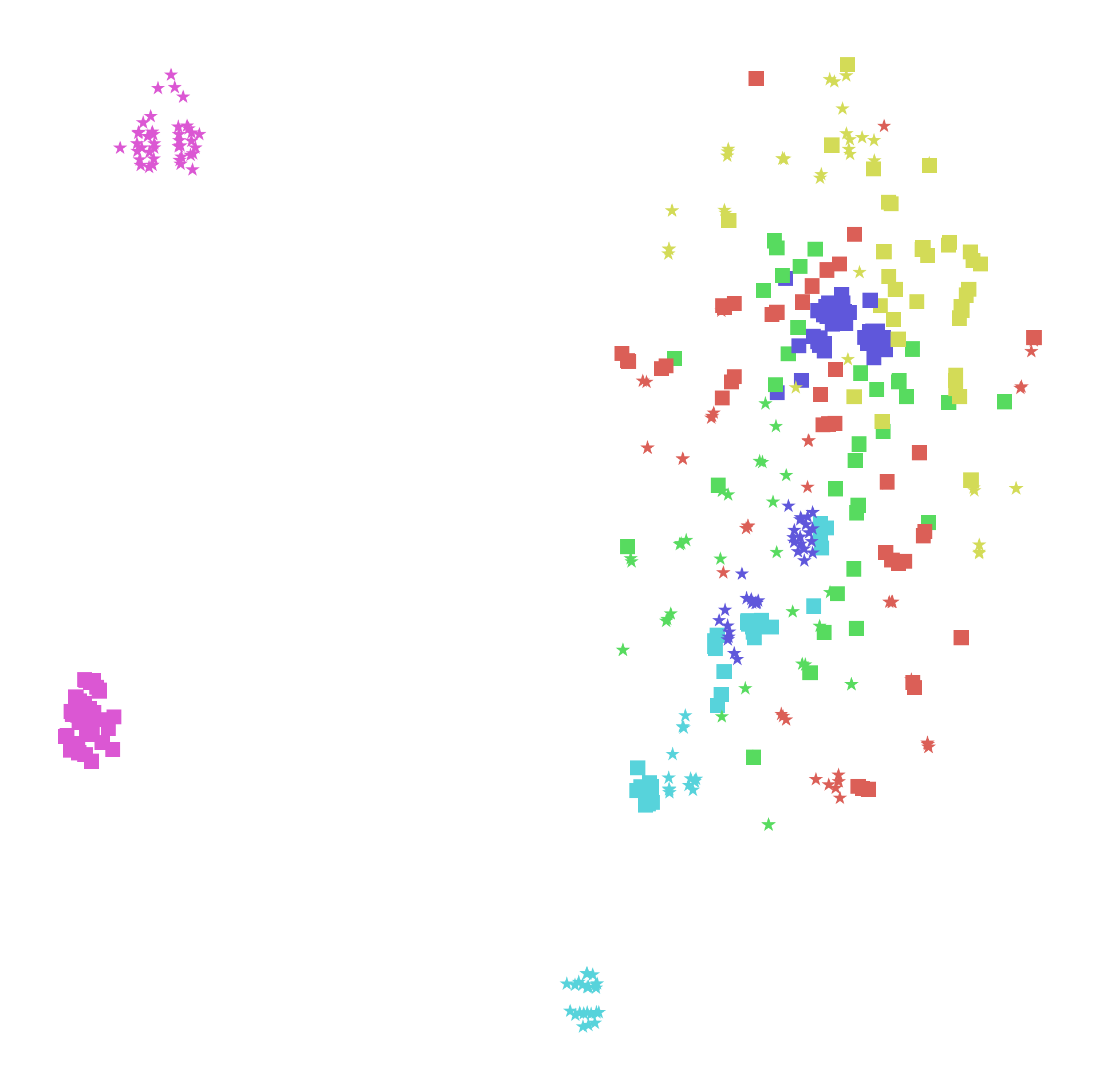}
    \label{fig:1a}
  }
  \subfigure[DeepCoral]{
    \includegraphics[width=0.21\textwidth]{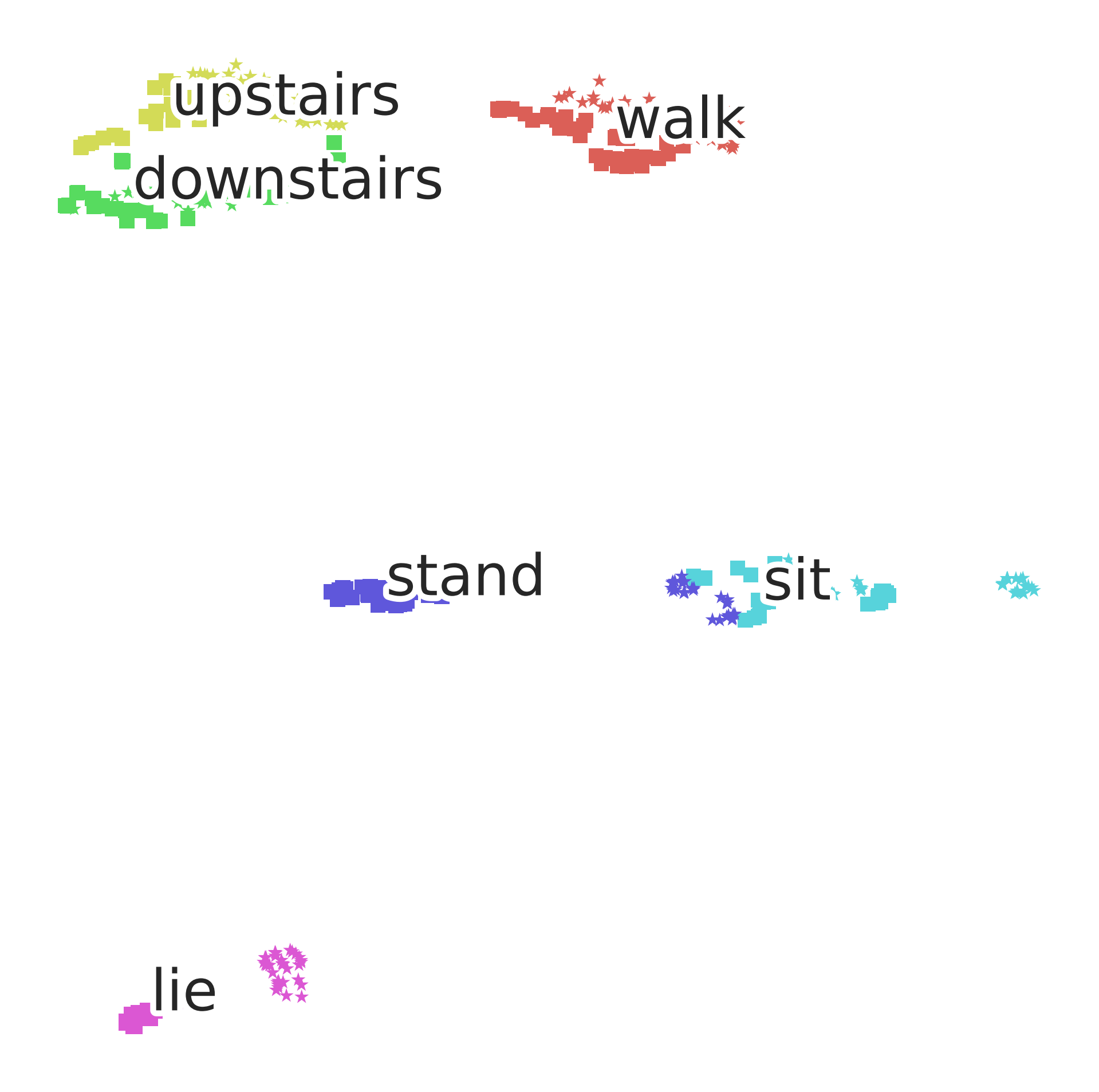}
    \label{fig:1b}
  }
  \subfigure[CLUDA]{
    \includegraphics[width=0.21\textwidth]{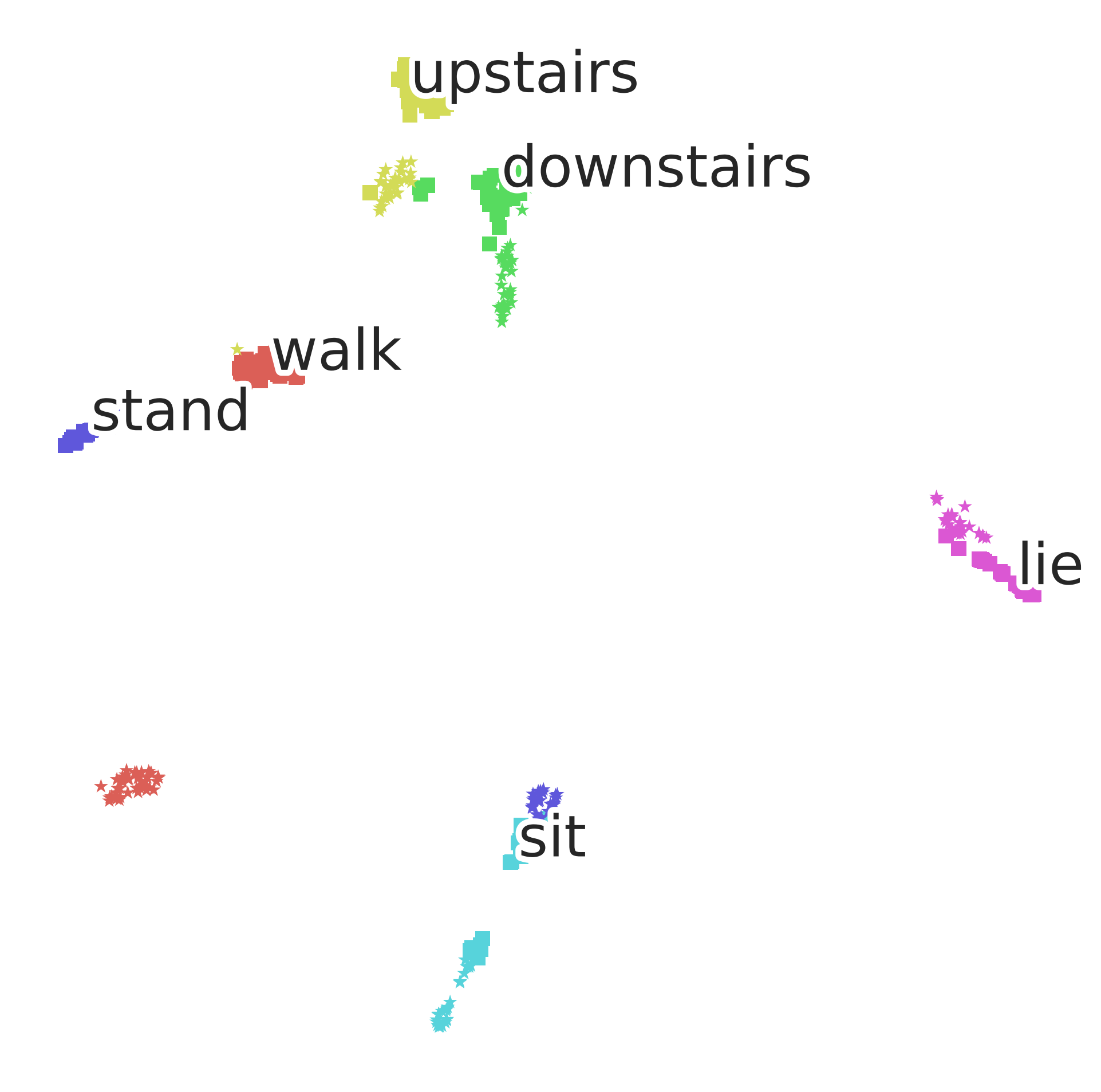}
    \label{fig:1c}
  }
  \subfigure[\methodname]{
    \includegraphics[width=0.21\textwidth]{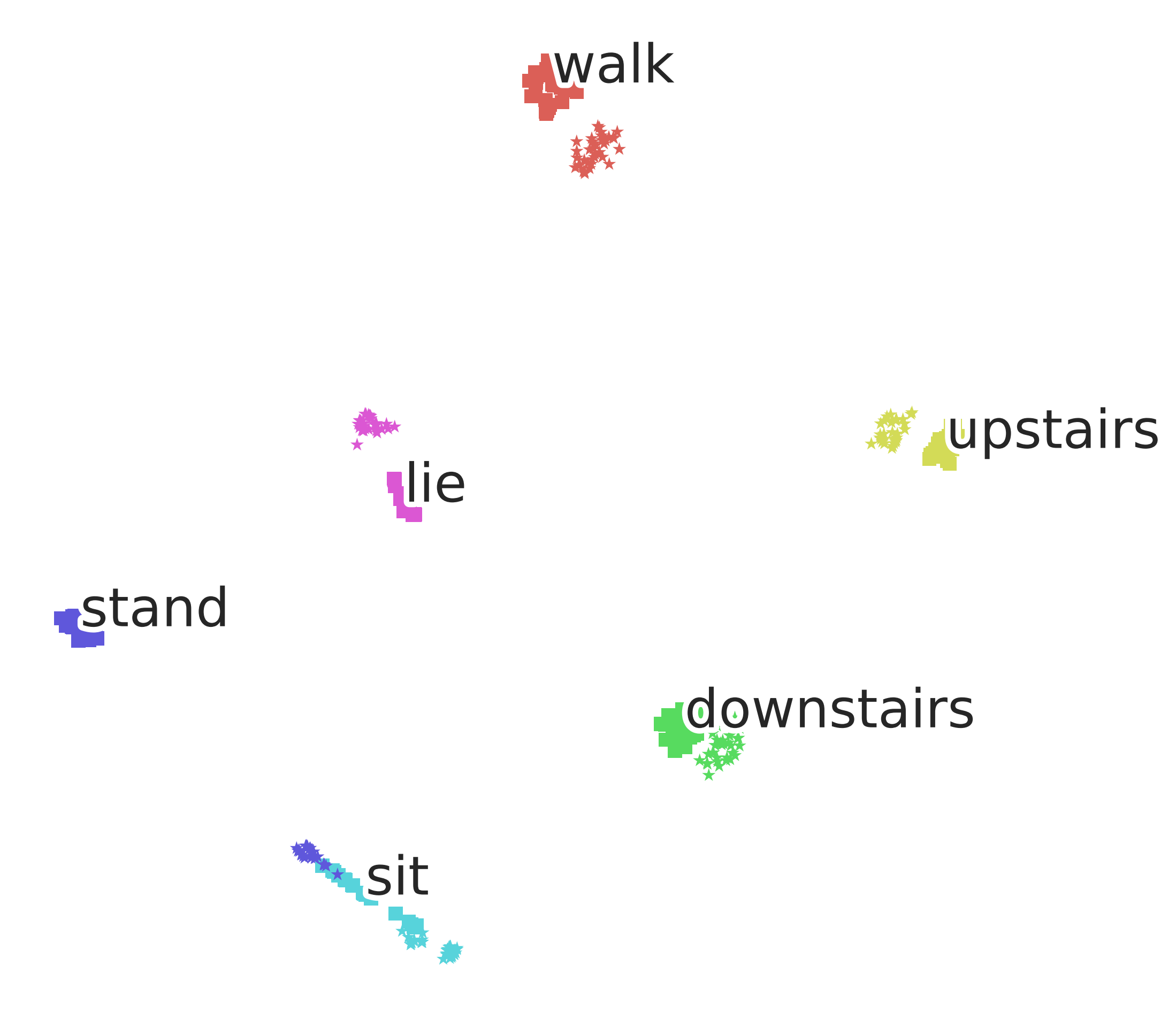}
    \label{fig:1d}
  }
  \caption{
  For the HAR dataset of adapting from source 2 to target 11, we generated T-SNE plots of learned embeddings for three different methods. Figure \ref{fig:1a} depicts the T-SNE visualization of datasets of source and target domains. Figure \ref{fig:1b}, \ref{fig:1c}, and \ref{fig:1d} represent a different method respectively, and the plots are arranged from left to right (DeepCoral, CLUDA, and \methodname respectively). In each plot, each color corresponds to a different activity label. The square markers represent embeddings of source samples, while the star markers represent embeddings of target samples. These T-SNE plots provide a visual representation of the learned embeddings and demonstrate the effectiveness of the different methods in adapting to the target domain.
  }
  \label{fig:tsne}
\end{figure}

\begin{figure}[htbp]
  \centering
  \subfigure[HAR 7- HAR 13]{
    \includegraphics[width=0.21\textwidth]{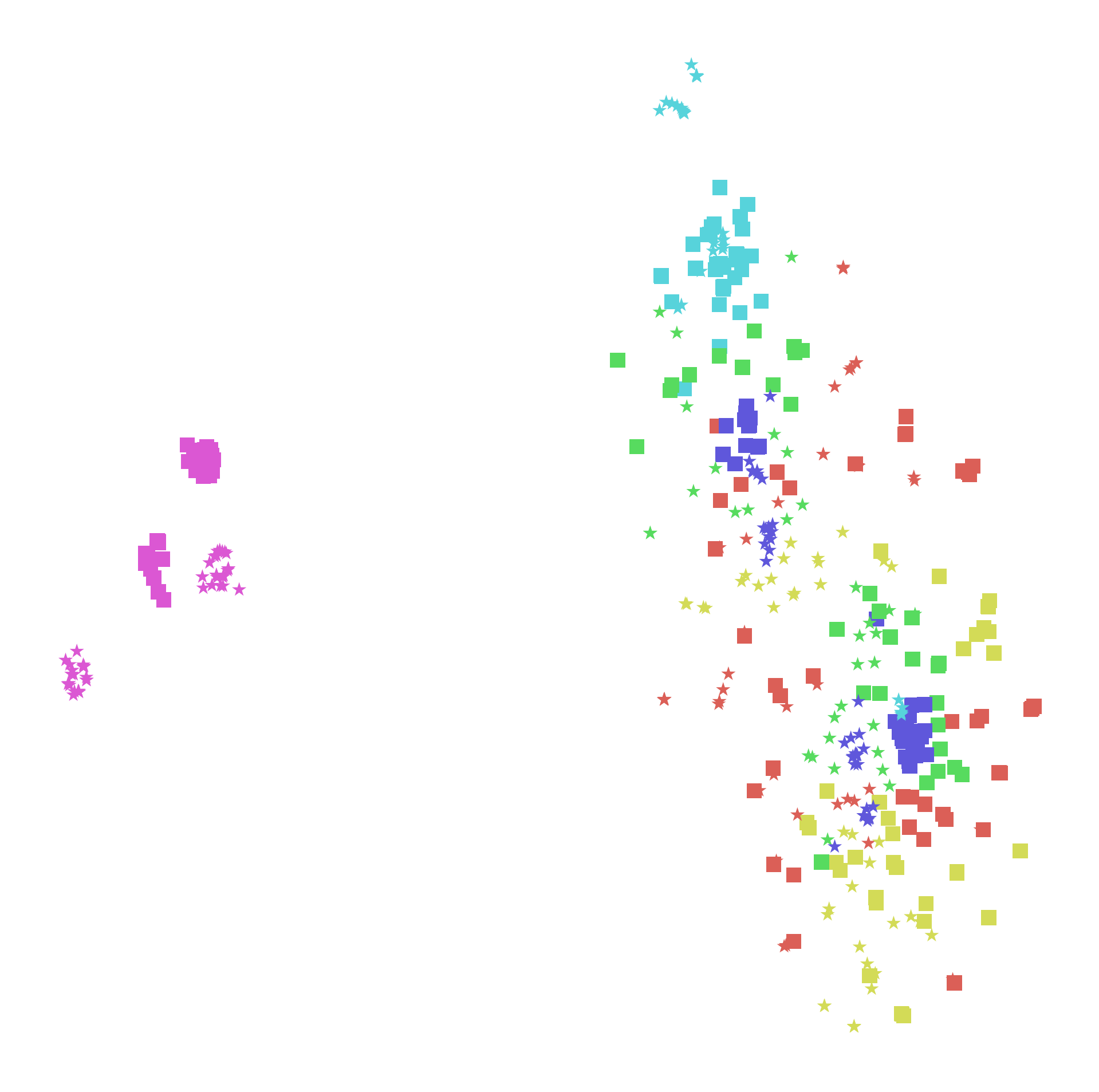}
    \label{fig:2a}
  }
  \subfigure[DeepCoral]{
    \includegraphics[width=0.21\textwidth]{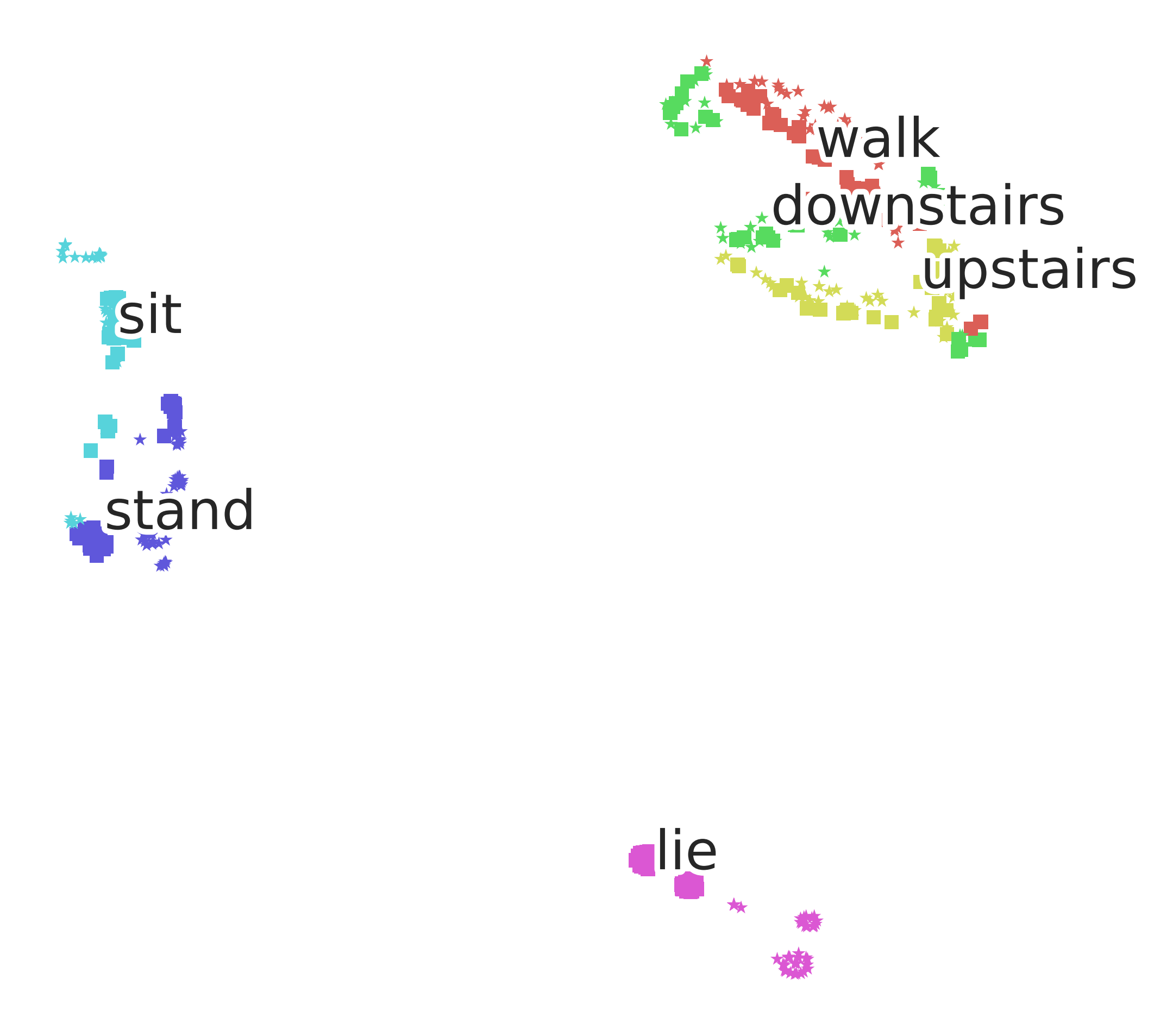}
    \label{fig:2b}
  }
  \subfigure[CLUDA]{
    \includegraphics[width=0.21\textwidth]{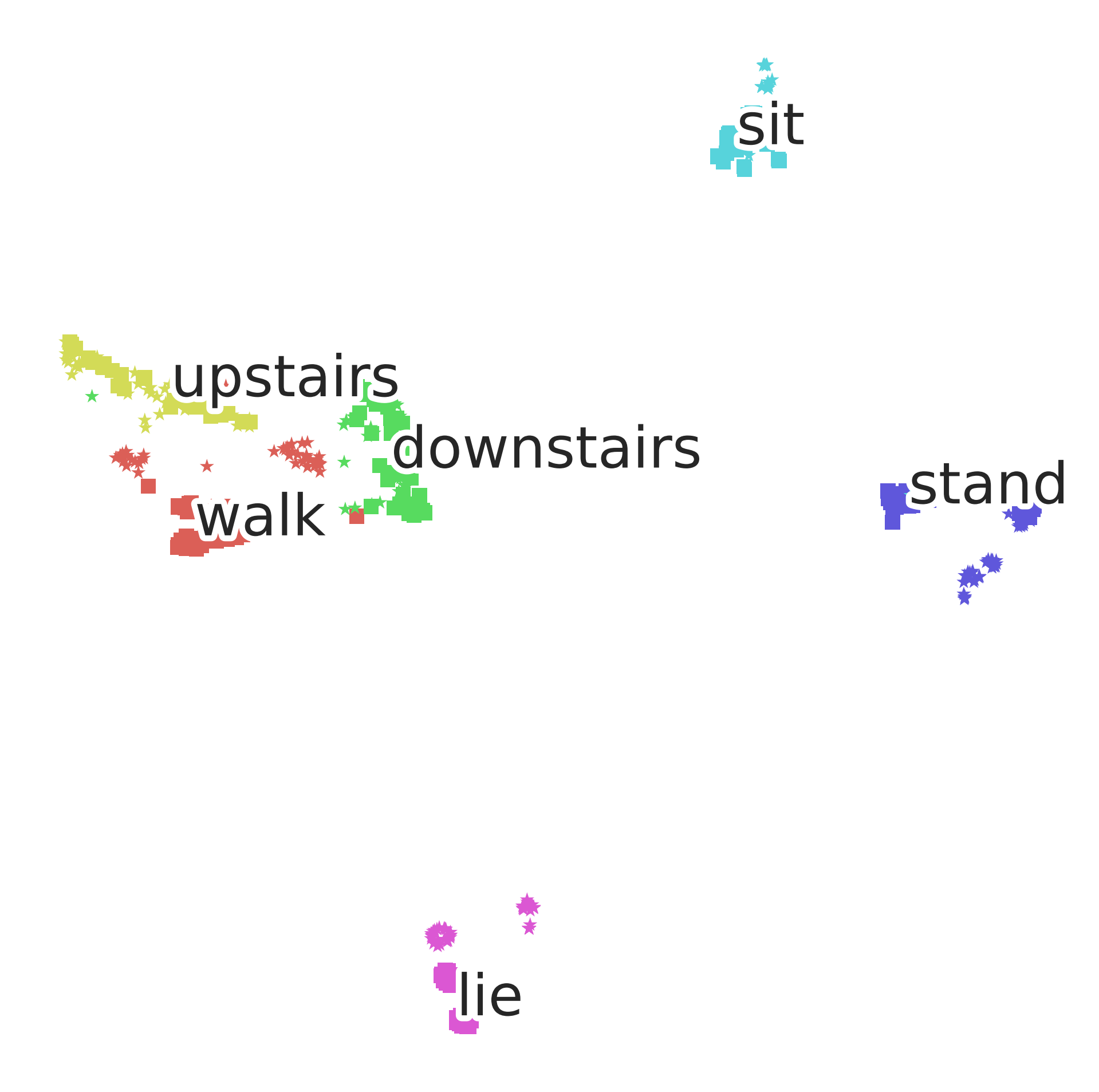}
    \label{fig:2c}
  }
  \subfigure[\methodname]{
    \includegraphics[width=0.21\textwidth]{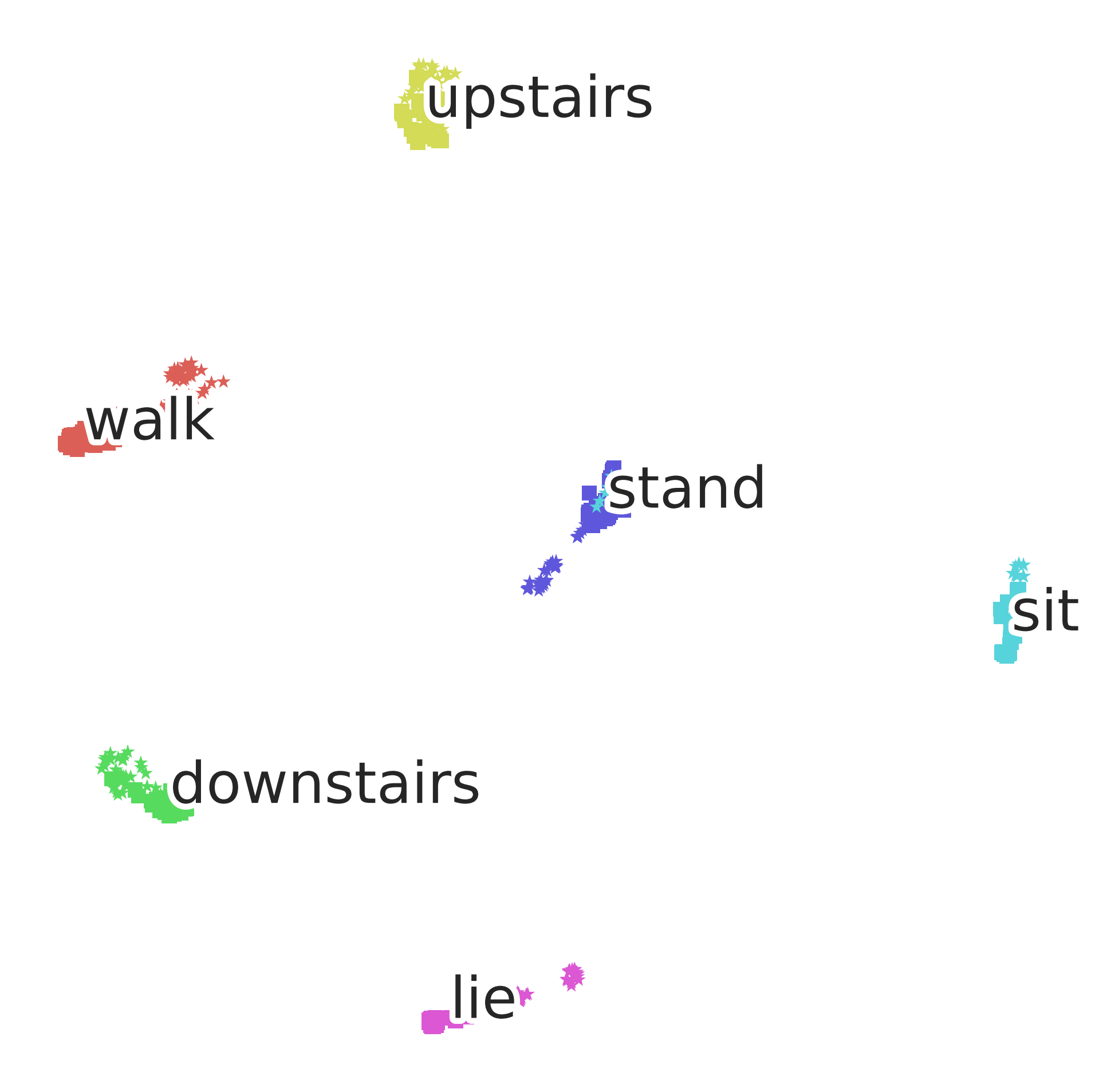}
    \label{fig:2d}
  }
  \caption{
  For the HAR dataset of adapting from source 7 to target 13, we generated T-SNE plots of learned embeddings for three different methods. In each plot, each color corresponds to a different activity label. The square markers represent source samples, while the star markers represent target samples. These T-SNE plots provide a visual representation of the learned embeddings and demonstrate the effectiveness of the different methods in adapting to the target domain.
  }
  \label{fig:tsne2}
\end{figure}

We present t-SNE plots of the learned representations using HAR in Figures \ref{fig:tsne} and \ref{fig:tsne2}, which serve as strong evidence of the effectiveness of \methodname for domain adaptation. The t-SNE plots provide visual representations of the feature distributions in both the source and target domains. 

The plots clearly depict distinct clusters of data points, corresponding to different activity types, in both the source and target domains. This observation indicates that \methodname successfully preserves the underlying structure of the data, even in the presence of differences in sensor configurations and other domain-specific factors. The distinct clusters in the t-SNE plots validate the ability of our method to capture and discriminate between different activities.

Additionally, the t-SNE plots reveal that the clusters in the target domain are generally more tightly grouped and better separated compared to those in the source domain. This suggests that \methodname effectively adapts the model to the target domain, leading to improved performance and more accurate predictions. These findings demonstrate the efficacy of \methodname for domain adaptation and highlight its potential for a wide range of applications, including robotics, healthcare, and sports performance analysis.

\subsection{Full Results of Closed-Set DA}
Comprehensive tables presenting the results for Closed-Set Domain Adaptation (DA) experiments are provided in two separate tables, accuracy and macro-F1. Table~\ref{tab:closed-acc} showcases the accuracy scores, while Table~\ref{tab:closed-f1} displays the Macro-F1 scores. Upon analyzing the tables, it becomes evident that \methodname consistently outperforms the baseline methods across all datasets in terms of both accuracy and Macro-F1 scores. This demonstrates the superiority of \methodname in effectively adapting to the target domain and achieving improved performance compared to the baseline approaches.

\begin{table*}[h!]
\scriptsize
  \caption{Prediction accuracy for each dataset between various subjects. Shown: mean Accuracy over 5 independent runs.}
  \label{tab:closed-acc}
  \centering
  \setlength{\tabcolsep}{2pt}
  \begin{tabular}{lccccccccc}
    \toprule
    Source $\mapsto$ Target & w/o UDA & CDAN  & DeepCORAL & {AdaMatch}  & DIRT-T & CLUDA & AdvSKM & CoDATS & \methodname  \\
   
    \midrule
    HAR 2 $\mapsto$ 11 & 76.56  & 85.42 & 90.63 & 75.00 & 80.21 & 81.77 & 98.96 & 68.23 &  \cellcolor{gray!50}\textbf{100}\\
    HAR 6 $\mapsto$ 23 & 67.36  & 87.50 & 84.38 & 80.20 & 74.31 & 92.01 & 88.54 & 74.31 & \cellcolor{gray!50}\textbf{95.83}\\
    HAR 7 $\mapsto$ 13 & 83.68  & 92.01 & 87.50 & 85.76 & 82.99 & 99.31 & 92.71 & 77.43 & \cellcolor{gray!50}\textbf{100}\\
    HAR 9 $\mapsto$ 18 & 24.65  & 58.86 & 46.88 & 56.59 & 59.03 & 67.71 &74.65 & 63.89 & \cellcolor{gray!50}\textbf{75.69}\\
    HAR 12 $\mapsto$ 16 & 61.11 & 66.67 & 65.28 & 49.65 & 67.01 & 65.28 & 69.44 & 66.32 & \cellcolor{gray!50}\textbf{86.52}\\
    HAR 13 $\mapsto$ 19 & 88.89 & 96.52 & 95.49 & 94.79 & 99.30 & 94.44 & 93.05 & 94.09 & \cellcolor{gray!50}\textbf{100}\\
    HAR 18 $\mapsto$ 21 & 100   & 100   & 100   & 100   & 98.61 & 98.96 & 100 & 99.65 & \cellcolor{gray!50}\textbf{100}\\
    HAR 20 $\mapsto$ 6  & 94.10 & 95.13 & 95.49 & 84.37 & 92.36 & \textbf{97.22} & 85.41 & 70.49 & \cellcolor{gray!50}93.41\\
    HAR 23 $\mapsto$ 13 & 71.18 & 82.64 & 69.79 & 68.75 & 74.72 & 72.92 & 79.51 & 56.25 & \cellcolor{gray!50}\textbf{86.52}\\
    HAR 24 $\mapsto$ 12 & 83.68 & 93.40 & 87.50 & 70.83 & 94.27 & \textbf{99.31} & 96.87 & 82.81 & \cellcolor{gray!50}93.75\\
    \hdashline
    HAR Avg  & 75.12 & 85.78 & 82.01 & 76.07 & 83.26 & 85.53 & 83.26 & 75.54 & \cellcolor{gray!50}\textbf{94.43}\\
    HAR Std of Avg & 0.98 & \textbf{0.91} & 1.09 & 1.77 & 2.78 & 1.78 & 2.79 & 3.31 & \cellcolor{gray!50}1.32\\
    \midrule

    HHAR 0 $\mapsto$ 2 & 64.51 & 76.19 & 84.23 & 84.78 & 77.83 & 79.84 & 78.94 & 79.61 & \cellcolor{gray!50}\textbf{87.72}\\
    HHAR 1 $\mapsto$ 6 & 70.63 & 92.57 & 90.14 & 92.31 & 88.54& \textbf{93.40} & 87.91 & 90.90 & \cellcolor{gray!50}93.33\\
    HHAR 2 $\mapsto$ 4 & 45.42 & 52.57 & 47.08 & 54.50 & 50.69& 45.90 & 52.57 & 60.07 & \cellcolor{gray!50}\textbf{63.75}\\
    HHAR 4 $\mapsto$ 0 & 32.81 & 29.09 & 28.13 & 36.45 & 32.22& 38.84 & 33.49 & 21.80 & \cellcolor{gray!50}\textbf{46.46}\\
    HHAR 4 $\mapsto$ 5 & 78.32 & 97.27 & 90.49 & 78.45 & 93.16& 94.08 & 92.64 & 97.66 & \cellcolor{gray!50}\textbf{98.05}\\
    HHAR 5 $\mapsto$ 1 & 90.63 & 96.16 & 89.91 & 94.20 & 91.86& 95.57 & 92.71 & 97.66 & \cellcolor{gray!50}\textbf{98.25}\\
    HHAR 5 $\mapsto$ 2 & 25.67 & 35.04 & 38.39 & 41.96 & 38.62& 33.93 & 36.53 & 41.44 & \cellcolor{gray!50}\textbf{42.63} \\
    HHAR 7 $\mapsto$ 2 & 32.37 & 37.05 & 34.45 & 37.65 & 38.10& 37.80 & 39.95 & 38.54 & \cellcolor{gray!50}\textbf{43.32}\\
    HHAR 7 $\mapsto$ 5 & 39.26 & 75.26 & 55.73 & 63.80 & 72.46& 75.26 & 65.49 & 58.15 & \cellcolor{gray!50}\textbf{84.17}\\
    HHAR 8 $\mapsto$ 4 & 62.92 & 96.11 & 76.88 & 64.69 & 65.83& 96.11 & 83.75 & \textbf{97.01} & \cellcolor{gray!50}93.75\\
   \hdashline
    HHAR Avg & 54.25 & 68.73 & 68.03 & 65.91 & 64.99 & 68.73 & 66.41 & 68.71 & \cellcolor{gray!50}\textbf{74.21}\\
    HHAR Std of Avg & 1.31 & 1.52 & 0.99 & 1.41 & 2.13 & 0.69 & \textbf{0.30} & 0.88 & \cellcolor{gray!50}0.72\\
     \midrule

    WISDM 2 $\mapsto$ 32 & 81.16 & \textbf{89.37} & 87.92 & 74.39 & 77.78 & 73.91 & 70.83 & 77.29 & \cellcolor{gray!50}79.71\\
    WISDM 4 $\mapsto$ 15 & 79.86 & 65.97 & 62.50 & 78.47 & 70.83 & 67.36 & 95.85 & 70.83 & \cellcolor{gray!50}\textbf{97.91}\\
    WISDM 7 $\mapsto$ 30 & 89.32 & 84.79 & 91.26 & 89.64 & 90.61 & 86.40 & 93.85 & 83.20 & \cellcolor{gray!50}\textbf{91.28}\\
    WISDM 12 $\mapsto$17 & 71.53 &  70.48 & 79.86 & 73.26 & 70.20 & 65.97 & 77.08 & 70.17 & \cellcolor{gray!50}\textbf{89.80}\\
    WISDM 12 $\mapsto$19 & 54.29 & 51.01 & 51.77 & 55.30 & 51.51 & 49.24 & 47.47 & 47.47 & \cellcolor{gray!50}\textbf{85.00}\\
    WISDM 18 $\mapsto$20 & 83.74 &  88.62 & 64.23 & 75.20 & 85.36 & 83.74 & 81.30 & 76.01 & \cellcolor{gray!50}\textbf{92.23}\\
    WISDM 20 $\mapsto$30 & 67.96 & 77.02 & 81.88 & 74.76 & 71.84 & 72.49 & 21.28 & 82.85 & \cellcolor{gray!50}\textbf{91.66}\\
    WISDM 21 $\mapsto$31 & 21.29 & 46.58 & 54.62 & 31.32 & 54.41 & 49.97 & 44.45 & 52.61 & \cellcolor{gray!50}\textbf{59.09}\\
    WISDM 25 $\mapsto$29 & 26.11 & 44.33 & 53.89 & 57.78 & 60.04 & 35.00 & 74.79 & 53.89 & \cellcolor{gray!50}\textbf{82.97}\\
    WISDM 26 $\mapsto$2  & 82.52 & 83.33 & 77.44 & 87.20 & 66.46 & \textbf{86.47} & 74.95 & 83.29 & \cellcolor{gray!50}83.50\\
\hdashline
    WISDM Avg & 65.78 & 70.05 & 70.80 & 69.79 & 69.62 & 67.04 & 66.97 & 70.66 & \cellcolor{gray!50}\textbf{76.60}\\
    WISDM Std of Avg & 1.92 & 1.01 & 1.16 & 1.01 & 1.41 & 0.91 & 1.84 & 0.88 & \cellcolor{gray!50}\textbf{0.73}\\
    \midrule

    Sleep-EDF 0 $\mapsto$ 11 & 55.60 & 68.94 & 57.22 & 63.86 &  65.88 & 57.87 &  56.51 &  69.53 &    \cellcolor{gray!50}\textbf{74.41}\\
    Sleep-EDF 2 $\mapsto$ 5 & 60.03 & 69.53 & 60.41 & 72.39 & 72.85 & 71.86 & 65.62   & 71.83 & \cellcolor{gray!50}\textbf{73.76}\\
    Sleep-EDF 12 $\mapsto$ 5 & 72.01 & 78.45  & 75.00 & 72.09 & 78.97 & 79.39 & 76.49 &  79.28 &  \cellcolor{gray!50}\textbf{79.81}\\
    Sleep-EDF 7 $\mapsto$ 18 & 53.91 &  73.18 & 65.82 & 71.61 & 74.34 & 74.49 &  60.93&  73.19 & \cellcolor{gray!50}\textbf{75.32}\\
    Sleep-EDF 16 $\mapsto$ 1 & 40.21 & 74.53  &  69.53 & 57.86 &  \textbf{81.82} & 75.83  & 72.96 & 75.32 & \cellcolor{gray!50}78.64 \\
    Sleep-EDF 9 $\mapsto$ 14 & 75.00  & 80.14& 82.22 &  82.55 & 86.14 & 86.32 &  76.75  & 81.64 & \cellcolor{gray!50}\textbf{87.17}\\
    Sleep-EDF 4 $\mapsto$ 12 & 48.76 & 67.08 & 64.97 & 48.17 & 68.48 & 66.53 &   66.14  & 71.68 & \cellcolor{gray!50}\textbf{69.86}\\
    Sleep-EDF 10 $\mapsto$ 7 & 67.86  & 74.35 & 76.05 & 60.41 & 75.05 & 75.23 & 74.31   & 73.31 & \cellcolor{gray!50}\textbf{77.23}\\
    Sleep-EDF 6 $\mapsto$ 3 & 75.20 & 80.99 & 78.38 & 78.12 & 83.66 & 81.96 &  78.90  & 83.59 & \cellcolor{gray!50}\textbf{84.58}\\
    Sleep-EDF 8 $\mapsto$ 10 & 35.21 & 55.16 & 36.79 & 51.25 & 46.01 & \textbf{65.70} & 44.76   & 44.22 & \cellcolor{gray!50}62.35\\
     \hdashline
    Sleep-EDF Avg & 58.38 & 72.24  & 66.66 & 65.83 & 66.04 & 73.50 & 67.33 & 72.36 & \cellcolor{gray!50}\textbf{76.31}\\
    Sleep-EDF Std of Avg & 1.33  & \textbf{0.54} & 1.16 & 1.69 & 0.99 & 0.34 & 0.89 & 1.03 & \cellcolor{gray!50}0.87\\
     \midrule
    Boiler 1 $\mapsto$ 2 & 57.09 & 67.93 & 67.13 & 67.42 &  68.13 & 68.93 & 72.43 & 75.74 & \cellcolor{gray!50}\textbf{98.06}\\
    Boiler 1 $\mapsto$ 3 & 74.54 & 94.98 & 93.32 & 94.02 &  94.88 & 95.36 & 96.14 & 97.32 & \cellcolor{gray!50}\textbf{99.57}\\
    Boiler 2 $\mapsto$ 1 & 73.14 & 85.96 & 84.32 & 84.32 &87.76 & 88.74& 89.32 & 90.23& \cellcolor{gray!50}\textbf{97.33}\\
    Boiler 2 $\mapsto$ 3 & 66.09 & 93.32 & 91.53& 92.89 & 92.62   & 91.31& 91.53 & 92.89 & \cellcolor{gray!50}\textbf{93.18}\\
    Boiler 3 $\mapsto$ 1 & 74.99 & 93.89 & 92.43 & 93.01 & 93.14 & 93.92 & 94.77 & 95.32 & \cellcolor{gray!50}\textbf{98.1}\\
    Boiler 3 $\mapsto$ 2 & 61.31 & 63.32 & 60.39 & 57.93 & 60.43  & 60.43 & 70.62 & 72.32 & \cellcolor{gray!50}\textbf{99.57}\\
     \hdashline
    Boiler Avg & 65.86 & 83.23 & 81.45 & 81.59 & 82.77 &  83.03 & 85.69 & 87.21& \cellcolor{gray!50}\textbf{97.64}\\
    Boiler Std of Avg & 0.84 & 1.02 & 0.73 & 0.78 & 0.81 & 0.97 & 0.64 & 0.69 & \cellcolor{gray!50}\textbf{0.51}\\
    \bottomrule
    \multicolumn{7}{l}{\scriptsize Higher is better. Best value in bold.}
  \end{tabular}
\end{table*}


\begin{table*}[h!]
\scriptsize
  \caption{Macro-F1 for each dataset between various subjects. Shown: mean Accuracy over 5 independent runs.}
  \label{tab:closed-f1}
  \centering
  \centering
  \setlength{\tabcolsep}{2pt}
  \begin{tabular}{lccccccccc}
    \toprule
    Source $\mapsto$ Target & w/o UDA & CDAN  & DeepCORAL & {AdaMatch}  & DIRT-T & CLUDA & AdvSKM & CoDATS & \methodname  \\
   
    \midrule
    HAR 2 $\mapsto$ 11 & 0.69 & 0.85 & 0.91 & 0.73 & 0.81 & 0.81 & 0.99 & 0.66 & \cellcolor{gray!50}\textbf{1.00}\\
    HAR 6 $\mapsto$ 23 & 0.63 & 0.88 & 0.81 & 0.81 & 0.68 & 0.92 & 0.87 & 0.71 & \cellcolor{gray!50}\textbf{0.96}\\
    HAR 7 $\mapsto$ 13 & 0.84 & 0.91 & 0.87 & 0.86 & 0.82 & 0.99 & 0.92 & 0.78 & \cellcolor{gray!50}\textbf{1.00}\\
    HAR 9 $\mapsto$ 18 & 0.17 & 0.61 & 0.44 & 0.55 & 0.58 & 0.67 & 0.73 & 0.60 & \cellcolor{gray!50}\textbf{0.76}\\
    HAR 12 $\mapsto$ 16 & 0.58 & 0.64 & 0.65 & 0.48 & 0.62 & 0.64 & 0.68 & 0.64 & \cellcolor{gray!50}\textbf{0.86}\\
    HAR 13 $\mapsto$ 19 & 0.91 & 0.97 & 0.95 & 0.94 & 0.99 & 0.94 & 0.93 & 0.93 & \cellcolor{gray!50}\textbf{1.00}\\
    HAR 18 $\mapsto$ 21 & 1.00 & 1.00 & 1.00 & 1.00 & 0.98 & 0.99 & 1.00 & 0.99 & \cellcolor{gray!50}\textbf{1.00}\\
    HAR 20 $\mapsto$ 6  & 0.94 & 0.95 & 0.95 & 0.84 & 0.92 & \textbf{0.98} & 0.84 & 0.65 & \cellcolor{gray!50}0.94\\
    HAR 23 $\mapsto$ 13 & 0.71 & 0.82 & 0.70 & 0.67 & 0.74 & 0.71 & 0.77 & 0.54 & \cellcolor{gray!50}\textbf{0.86}\\
    HAR 24 $\mapsto$ 12 & 0.84 & 0.92 & 0.88 & 0.70 & 0.93 & \textbf{0.99} & 0.96 & 0.81 & \cellcolor{gray!50}0.94\\
    \hdashline
    HAR Avg  & 0.73 & 0.86 & 0.82 & 0.76 & 0.81 & 0.86 & 0.87 & 0.72 &\cellcolor{gray!50}\textbf{0.93}\\
    HAR Std of Avg  & 0.024 & 0.014 & 0.015 & 0.011 & 0.032 & 0.005 & 0.010 & 0.04 & \cellcolor{gray!50}\textbf{0.005}\\
    \midrule

    HHAR 0 $\mapsto$ 2 & 0.60 & 0.70 & 0.86 & 0.83 & 0.76 & 0.82 & 0.72 & 0.73 &\cellcolor{gray!50}\textbf{0.87}\\
    HHAR 1 $\mapsto$ 6 & 0.64 & 0.93 & 0.91 & 0.93 & 0.86 & \textbf{0.94} & 0.88 & 0.90&\cellcolor{gray!50}0.93\\
    HHAR 2 $\mapsto$ 4 & 0.32 & 0.52 & 0.45 & 0.46 & 0.51 & 0.44 & 0.44 & 0.46&\cellcolor{gray!50}\textbf{0.59}\\
    HHAR 4 $\mapsto$ 0 & 0.29 & 0.27 & 0.26 & 0.32 & 0.30 & 0.40 & 0.33& 0.20&\cellcolor{gray!50}\textbf{0.45}\\
    HHAR 4 $\mapsto$ 5 & 0.78 & 0.98 & 0.90 & 0.76 & 0.93 & 0.94 & 0.93& 0.96&\cellcolor{gray!50}\textbf{0.98}\\
    HHAR 5 $\mapsto$ 1 & 0.90 & 0.98 & 0.90 & 0.94 & 0.90 & 0.96 & 0.92& 0.94&\cellcolor{gray!50}\textbf{0.98}\\
    HHAR 5 $\mapsto$ 2 & 0.19 & 0.35 & 0.36 & 0.40 & 0.36 & 0.37 & 0.35& 0.41&\cellcolor{gray!50}\textbf{0.41}\\
    HHAR 7 $\mapsto$ 2 & 0.31 & 0.32 & 0.32 & 0.37 & 0.34 & 0.36 & 0.41& 0.36&\cellcolor{gray!50}\textbf{0.44}\\
    HHAR 7 $\mapsto$ 5 & 0.36 & 0.76 & 0.50 & 0.60 & 0.73 & 0.65 & 0.64& 0.59&\cellcolor{gray!50}\textbf{0.86}\\
    HHAR 8 $\mapsto$ 4 & 0.58 & 0.97 & 0.73 & 0.61 & 0.64 & 0.84 & 0.83& \textbf{0.95}&\cellcolor{gray!50}0.94\\
   \hdashline
    HHAR Avg & 0.5 & 0.68 & 0.62 & 0.62 & 0.64 & 0.67 & 0.65 & 0.63 & \cellcolor{gray!50}\textbf{0.75}\\
    HHAR Std of Avg & 0.022 & 0.013 & 0.007 & 0.013 & 0.023 & 0.008 & \textbf{0.003} & 0.006 &\cellcolor{gray!50} 0.004\\
     \midrule

    WISDM 2 $\mapsto$ 32 & 0.68 & \textbf{0.72} & 0.71 & 0.59 & 0.65 & 0.64 & 0.61 & 0.66 & 
    \cellcolor{gray!50} 0.68\\
    WISDM 4 $\mapsto$ 15 & 0.52 & 0.44 & 0.42 & 0.54 & 0.41 & 0.61 & 0.55 & 0.41 & 
    \cellcolor{gray!50} \textbf{0.98}\\
    WISDM 7 $\mapsto$ 30 & 0.77 & 0.70 & 0.85 & 0.76 & 0.78 & 0.81 & 0.84 & 0.75& 
    \cellcolor{gray!50} \textbf{0.86}\\
    WISDM 12 $\mapsto$17 & 0.53 & 0.50 & 0.67 & 0.67 & 0.56 & 0.59 & 0.53& 0.62 &  \cellcolor{gray!50} \textbf{0.72}\\
    WISDM 12 $\mapsto$19 & 0.36 & 0.31 & 0.35 & 0.38 & 0.39 & 0.41 & 0.35& 0.37 &  \cellcolor{gray!50} \textbf{0.78}\\
    WISDM 18 $\mapsto$20 & 0.81 & 0.87 & 0.63 & 0.66 & 0.67 & 0.70 & 0.71& 0.76&
     \cellcolor{gray!50} \textbf{0.92}\\
    WISDM 20 $\mapsto$30 & 0.56 & 0.64 & 0.67 & 0.54 & 0.65 & 0.70 & 0.61& 0.72& \cellcolor{gray!50} \textbf{0.87}\\
    WISDM 21 $\mapsto$31 & 0.10 & 0.31 & 0.27 & 0.16 & 0.28 & 0.27 & 0.28 & 0.30 &\cellcolor{gray!50} \textbf{0.43}\\
    WISDM 25 $\mapsto$29 & 0.15 & 0.23 & 0.25 & 0.24 & 0.21 & 0.26 & 0.28& 0.30 &\cellcolor{gray!50} \textbf{0.44}\\
    WISDM 26 $\mapsto$2  & 0.69 & 0.71 & 0.64 & 0.74 & 0.54 & 0.75 & 0.55 & 0.70& \cellcolor{gray!50} \textbf{0.75}\\
    \hdashline
    WISDM Avg & 0.52 & 0.54 & 0.52 & 0.54 & 0.54 & 0.57 & 0.55 & 0.56 & \cellcolor{gray!50} \textbf{0.74}\\
    WISDM Std of Avg & 0.031 & 0.020 & \textbf{0.006} & 0.015 & 0.012 & 0.029 &0.013 & 0.014 & \cellcolor{gray!50} 0.010\\
    \midrule

    Sleep-EDF 0 $\mapsto$ 11 & 0.48 & 0.54 & 0.50 & 0.52 & 0.53 & 0.47 &  0.48 & 0.50 &  \cellcolor{gray!50}\textbf{0.54}\\
    Sleep-EDF 2 $\mapsto$ 5 & 0.47 & 0.62 & 0.53 &  0.62 & 0.63 & 0.66 & 0.59   & 0.53 & \cellcolor{gray!50}\textbf{0.65}\\
    Sleep-EDF 12 $\mapsto$ 5 & 0.59 & 0.68  & 0.65 & 0.66 & 0.67 & 0.69 &  0.64&  0.66 &  \cellcolor{gray!50}\textbf{0.70}\\
    Sleep-EDF 7 $\mapsto$ 18 & 0.53 &  0.69 &0.62 & 0.59 & 0.71 & 0.71 &  0.60&  0.61 & \cellcolor{gray!50}\textbf{0.72}\\
    Sleep-EDF 16 $\mapsto$ 1 & 0.43 & 0.62 & 0.58 & 0.48 & 0.66 & 0.67  &  0.63 & 0.58 & \cellcolor{gray!50}\textbf{0.70}\\
    Sleep-EDF 9 $\mapsto$ 14 & 0.61  & 0.68 & 0.71 & 0.67 & 0.75 & 0.72&  0.68   & 0.71 & \cellcolor{gray!50}\textbf{0.76} \\
    Sleep-EDF 4 $\mapsto$ 12 & 0.42 & 0.59 & 0.59 &  0.37 & 0.59 &0.55 &   0.59  & 0.58 & \cellcolor{gray!50}\textbf{0.62}\\
    Sleep-EDF 10 $\mapsto$ 7 & 0.58  & 0.67 & 0.72 &  0.37 & 0.68  & 0.71&  0.72   & 0.71 & \cellcolor{gray!50}\textbf{0.73}\\
    Sleep-EDF 6 $\mapsto$ 3 & 0.67 & 0.73 &  0.70 & 0.62 & 0.75 &  0.73&   0.72  & 0.70 & \cellcolor{gray!50}\textbf{0.75}\\
    Sleep-EDF 8 $\mapsto$ 10 & 0.41 & 0.43 & 0.36 &  0.46 & 0.39 &  \textbf{0.65} & 0.46   & 0.38 & \cellcolor{gray!50}0.61\\
     \hdashline
    Sleep-EDF Avg & 0.52 & 0.63  & 0.60 & 0.54 & 0.64 & 0.65 & 0.61 & 0.60 & \cellcolor{gray!50}\textbf{0.68}\\
    Sleep-EDF Std of Avg & 0.026  & 0.005 & 0.015 & 0.004 & 0.005 & 0.007 & \textbf{0.003} & 0.012 & \cellcolor{gray!50}0.008\\
     \midrule
    Boiler 1 $\mapsto$ 2 & 0.52 & 0.63 & 0.63 & 0.64 &  0.65 & 0.68 & 0.73 & 0.73 & \cellcolor{gray!50}\textbf{0.98}\\
    Boiler 1 $\mapsto$ 3 & 0.74 & 0.95 & 0.93 & 0.94 &  0.95 & 0.95 & 0.96 & 0.97 & \cellcolor{gray!50}\textbf{0.98}\\
    Boiler 2 $\mapsto$ 1 & 0.70 & 0.81 & 0.83 & 0.83 & 0.85 & 0.86 & 0.88 & 0.91 & \cellcolor{gray!50}\textbf{0.97}\\
    Boiler 2 $\mapsto$ 3 & 0.60 & 0.91 & 0.90 & 0.91 & 0.91   & 0.90 & 0.90 & 0.91 & \cellcolor{gray!50}\textbf{0.91}\\
    Boiler 3 $\mapsto$ 1 & 0.70 & 0.94 & 0.90 & 0.93 & 0.92 & 0.94 & 0.94 & 0.95 & \cellcolor{gray!50}\textbf{0.97}\\
    Boiler 3 $\mapsto$ 2 & 0.55 & 0.59 & 0.60 & 0.54 & 0.61  & 0.58 & 0.69 & 0.70 & \cellcolor{gray!50}\textbf{0.99}\\
     \hdashline
    Boiler Avg & 0.635 & 0.80 & 0.80 & 0.80 & 0.82 & 0.82 &  0.85 & 0.86 & \cellcolor{gray!50}\textbf{0.97}\\
    Boiler Std of Avg & 0.008 & 0.010 & 0.007 & 0.008 & 0.010 & 0.006 & 0.007 & 0.005 & \cellcolor{gray!50}\textbf{0.005} \\
    \bottomrule
    \multicolumn{7}{l}{\scriptsize Higher value indicates better performance. Best value in bold.}
  \end{tabular}
\end{table*}

\subsection{Full Results of UniDA}
Detailed tables containing the results for Universal Domain Adaptation (UniDA) experiments are provided in two separate tables, accuracy and H-scores. Table~\ref{tab:uni-acc} presents the accuracy scores, while Table~\ref{tab:appendixUni-hscore} displays the H-scores. It is worth noting that accuracy alone is not an appropriate metric for evaluating UniDA since it does not fully reflect the ability to detect target unknown samples. Accuracy can be misleading due to class imbalance issues, resulting in high or low scores without effectively capturing the capability of detecting unknown samples. We conducts three UniDA  settings including WISDM$\to$WISDM, WISDM$\to$HHAR, HHAR$\to$WISDM. It can be observed that \methodname consistently outperforms the baseline methods across all three UniDA settings considered in this work. The superiority of \methodname is demonstrated in its ability to effectively handle the challenges associated with Universal Domain Adaptation and achieve improved performance compared to the baseline approaches.

\begin{table*}[h!]
\scriptsize
  \caption{Accuracy of UniDA using WISDM, WISDM$\to$HHAR, HHAR$\to$WISDM, Shown: mean Accuracy over 5 independent runs. Closed-Set DA baselines are colored in \textcolor{blue}{blue}. }
  \label{tab:uni-acc}
  \centering
  \setlength{\tabcolsep}{2pt}
  \begin{tabular}{lcccccccc}
    \toprule
    Source $\mapsto$ Target & No. Tar Private Class & \cellcolor{blue!20}CLUDA & UAN & DANCE  & OVANet & UniOT & \cellcolor{blue!20}\methodname (A) & \methodname (with A\&C)  \\

    \midrule
    WISDM 3 $\mapsto$ 2 & 1 & 31.71 & 8.04 & 8.53 &  25.61 & 26.78  & 28.05 & \cellcolor{gray!50}\textbf{28.05}\\
    WISDM 3 $\mapsto$ 7 & 1 & 23.96 & 8.19 & 8.33 &  \textbf{34.38} & 30.31 & 25.92  & \cellcolor{gray!50}25.92\\
    WISDM 13 $\mapsto$ 15 & 2 & 54.58 & 9.85 & 14.58 &  10.42  & 16.46& 58.33 & \cellcolor{gray!50}\textbf{64.58}\\
    WISDM 14 $\mapsto$ 19 & 2 & 30.30 & 39.03 & 44.00 &  42.42  & 40.32 & 46.21 & \cellcolor{gray!50}\textbf{53.78}\\
    WISDM 27 $\mapsto$ 28 & 2 & 8.98 & 6.94 & 6.74 & 7.87  & 10.98 & 22.92 & \cellcolor{gray!50}\textbf{53.70}\\
    WISDM 1 $\mapsto$ 0 & 2 & 71.05 & 70.34 & 75.71 & 74.29  & 73.14 & 73.68 & \cellcolor{gray!50}\textbf{82.57}\\
    WISDM 1 $\mapsto$ 3 & 3 & 0.00 & 32.85 & 38.46 &  \textbf{61.54} & 36.31 & 11.54 & \cellcolor{gray!50}35.54\\
    WISDM 10 $\mapsto$ 11 & 4 & 60.52 & 31.80 & 30.26 &  35.53 & 39.35& 72.37 & \cellcolor{gray!50}\textbf{76.36}\\
    WISDM 22 $\mapsto$ 17 & 4 & 26.32 & 27.87 & 23.68 & 40.79 & 38.31& 40.79 & \cellcolor{gray!50}\textbf{48.16}\\
    WISDM 27 $\mapsto$ 15 & 4 & 56.25 & 22.18 & 27.08 & 60.42 & 52.34& 58.17 & \cellcolor{gray!50}\textbf{66.42}\\

    \hdashline
    WISDM Avg  & & 36.37 & 25.71 & 27.70 & 33.28 & 36.43 & 44.08 & \cellcolor{gray!50}\textbf{53.51}\\
    WISDM Std of Avg & & 1.05 & 2.09 & 1.95 &   \textbf{0.97} & 1.25 & 1.06 & \cellcolor{gray!50}1.41\\
    \midrule
    W$\to$H 4 $\mapsto$ 0 & 1 &   32.43  & 24.5 & 30.73 & 35.24 & 36.51& 34.32  & \cellcolor{gray!50}\textbf{44.14} \\
    W$\to$H 5 $\mapsto$ 1 & 1 &   20.32  & 31.0 & 15.32 & 26.31& 28.14 &  27.94  & \cellcolor{gray!50}\textbf{35.65} \\
    W$\to$H 6 $\mapsto$ 2 & 1 &   60.32  & 34.7 & 32.32 & 40.35& 48.94&  65.12  & \cellcolor{gray!50}\textbf{69.01} \\
    W$\to$H 7 $\mapsto$ 3 & 1 &   51.84  & 21.10& 36.84 & 39.46 & 50.35& 55.10  & \cellcolor{gray!50}\textbf{60.88}  \\
    W$\to$H 17 $\mapsto$ 4 & 1 &  12.31  & 24.50 & 15.94 & 25.31& 26.32& 24.98  & \cellcolor{gray!50}\textbf{28.41} \\
    W$\to$H 18 $\mapsto$ 5 & 1 &  35.85  & 26.60 & 29.65& 36.14& 33.46& 35.70  & \cellcolor{gray!50}\textbf{40.76}  \\
    W$\to$H 19 $\mapsto$ 6 & 1 &  46.39  &32.75 & 38.13& 47.98& 49.32&  50.17 &  \cellcolor{gray!50}\textbf{54.76}  \\
    W$\to$H 20 $\mapsto$ 7 & 1 &  62.32  & 39.83 & 42.90& 58.11& 60.31&  64.98 &  \cellcolor{gray!50}\textbf{64.98}  \\
    W$\to$H 23 $\mapsto$ 8 & 1 &  53.76  & 32.71 & 40.87& 58.32 & 52.47&  60.71 & \cellcolor{gray!50}\textbf{62.84}   \\
    \hdashline
    W$\to$H Avg & & 37.55 & 29.74 & 39.06 & 40.80& 42.87& 46.55 & \cellcolor{gray!50}\textbf{51.35}\\
    W$\to$H Std of Avg & & \textbf{1.04} & 1.38 & 1.98 & 1.65& 1.74& 1.31 & \cellcolor{gray!50}1.22\\
     \midrule
    H$\to$W 0 $\mapsto$ 4 & 1 & 59.32 & 55.30 & 61.94 & 63.14 & 64.07 & 62.98 & \cellcolor{gray!50}\textbf{64.84}\\
    H$\to$W 1 $\mapsto$ 5 & 1 & 56.17 & 50.33 & 58.10 & 60.14 & 61.46& 60.94 &  \cellcolor{gray!50}\textbf{62.85}\\
    H$\to$W 2 $\mapsto$ 6 & 1 & 50.44 & 49.85 & 52.51 & 54.84 & 56.15& 55.95 & \cellcolor{gray!50}\textbf{57.11}\\
    H$\to$W 3 $\mapsto$ 7 & 1 & 52.21 & 53.01 & 55.91 & 55.71& 58.91 & 56.42 & \cellcolor{gray!50}\textbf{60.95}\\
    H$\to$W 4 $\mapsto$ 17 & 1 & 39.87 & 37.04 & 41.39 & 41.01& 42.50& 41.94 & \cellcolor{gray!50}\textbf{44.95}\\
    H$\to$W 5 $\mapsto$ 18 & 1 & 47.72 & 47.80 & 50.35 & 51.87& 52.22& 49.95 & \cellcolor{gray!50}\textbf{51.27} \\
    H$\to$W 6 $\mapsto$ 19 & 1 & 44.50 & 43.09 & 46.19 & 44.08& 45.93& 46.05 & \cellcolor{gray!50}\textbf{51.86}\\
    H$\to$W 7 $\mapsto$ 20 & 1 & 50.92 & 54.01 & 59.85 & 61.35 &61.06 & 47.00 & \cellcolor{gray!50}\textbf{62.59}\\
    H$\to$W 8 $\mapsto$ 23 & 1 & 44.50 & 42.06 & 43.66 &48.14 & 49.71& 47.77 & \cellcolor{gray!50}\textbf{52.64}\\
    \hdashline
    H$\to$W Avg & & 44.47 & 48.05 & 52.22 & 53.36 & 54.67 & 52.11 & \cellcolor{gray!50}\textbf{56.57} \\
    H$\to$W Std of Avg & & 1.31 & 1.39 & 1.21 & 0.94 & 1.05& \textbf{0.97} & \cellcolor{gray!50}1.08\\
    \bottomrule
    \multicolumn{5}{l}{\scriptsize Higher accuracy is better. Best value in bold.}
  \end{tabular}
  \label{tab:UniDA-acc}
\end{table*}

\begin{table*}[h!]
\scriptsize
  \caption{H-Score of UniDA using WISDM, WISDM$\to$HHAR, HHAR$\to$WISDM, Shown: mean Accuracy over 5 independent runs.}
  \label{tab:appendixUni-hscore}
  \centering
  \setlength{\tabcolsep}{2pt}
  \begin{tabular}{lcccccc}
    \toprule
    Source $\mapsto$ Target & No. Tar Private Class  & UAN & DANCE  & OVANet & UniOT  & \methodname   \\

    \midrule
    WISDM 3 $\mapsto$ 2 & 1 & 0 & 0 & 0.07 & 0.11 & \cellcolor{gray!50}\textbf{0.51}\\
    WISDM 3 $\mapsto$ 7 & 1 & 0& 0 & 0.2 & 0.22 & \cellcolor{gray!50}\textbf{0.52}  \\
    WISDM 13 $\mapsto$ 15 & 2  & 0& 0.14 & 0.33 & 0.36 & \cellcolor{gray!50}\textbf{0.50}\\
    WISDM 14 $\mapsto$ 19 & 2 & 0.24& 0.28 & 0.31 & 0.28 & \cellcolor{gray!50}\textbf{0.55}\\
    WISDM 27 $\mapsto$ 28 & 2 & 0.07& 0.07 & 0.23 & 0.35 & \cellcolor{gray!50}\textbf{0.59} \\
    WISDM 1 $\mapsto$ 0 & 2 & 0.41 & 0.39 & 0.38  & 0.40 & \cellcolor{gray!50}\textbf{0.43}\\
    WISDM 1 $\mapsto$ 3 & 3 & 0.46 &0.49 & 0.45 & 0.43 & \cellcolor{gray!50}\textbf{0.51}\\
    WISDM 10 $\mapsto$ 11 & 4  & 0 & 0 & 0.34 & 0.41 & \cellcolor{gray!50}\textbf{0.53}\\
    WISDM 22 $\mapsto$ 17 & 4 & 0.13 & 0 & 0.32 & 0.41& \cellcolor{gray!50}\textbf{0.52}\\
    WISDM 27 $\mapsto$ 15 & 4 & 0.43 & 0.51 & 0.46 & 0.52 & \cellcolor{gray!50}\textbf{0.57}\\

    \hdashline
    WISDM Avg  &  & 0.17 & 0.19 & 0.31 & 0.35& \cellcolor{gray!50}\textbf{0.52}\\
    WISDM Std of Avg & & 0.04 & 0.05 & 0.04& 0.05& \cellcolor{gray!50}\textbf{0.04}\\
    \midrule
    W$\to$H 4 $\mapsto$ 0 & 1 & 0 & 0.14 & 0.15 & 0.19& \cellcolor{gray!50}\textbf{0.49}\\
    W$\to$H 5 $\mapsto$ 1 & 1 & 0.24 & 0.22 & 0.25 & 0.28 & \cellcolor{gray!50}\textbf{0.53}\\
    W$\to$H 6 $\mapsto$ 2 & 1 & 0.14 & 0.12 & 0.20 & 0.25 & \cellcolor{gray!50}\textbf{0.55}\\
    W$\to$H 7 $\mapsto$ 3 & 1 & 0 & 0.15 & 0.04 & 0.14 & \cellcolor{gray!50}\textbf{0.51}\\
    W$\to$H 17 $\mapsto$ 4 & 1 & 0.35 & 0.28 & 0.41 & 0.45& \cellcolor{gray!50}\textbf{0.57}\\
    W$\to$H 18 $\mapsto$ 5 & 1 & 0.20 & 0.27 & 0.29 & 0.32& \cellcolor{gray!50}\textbf{0.47}\\
    W$\to$H 19 $\mapsto$ 6 & 1 & 0.19 & 0.22 & 0.25 & 0.28& \cellcolor{gray!50}\textbf{0.51}\\
    W$\to$H 20 $\mapsto$ 7 & 1 & 0.11 & 0.17 & 0.35 & 0.41& \cellcolor{gray!50}\textbf{0.49}\\
    W$\to$H 23 $\mapsto$ 8 & 1 & 0.21 & 0.28 & 0.47& 0.51& \cellcolor{gray!50}\textbf{0.57}\\
    \hdashline
    W$\to$H Avg & & 0.16 & 0.21 & 0.24 & 0.28 & \cellcolor{gray!50}\textbf{0.52} \\
    W$\to$H Std of Avg & & 0.03 & 0.02 & 0.03 & 0.02 & \cellcolor{gray!50}\textbf{0.02}\\
     \midrule
    H$\to$W 0 $\mapsto$ 4 & 1 & 0.23 & 0.28 & 0.33 & 0.37 & \cellcolor{gray!50}\textbf{0.45}\\
    H$\to$W 1 $\mapsto$ 5 & 1 & 0.19 & 0.31 & 0.38 & 0.42 & \cellcolor{gray!50}\textbf{0.47}\\
    H$\to$W 2 $\mapsto$ 6 & 1 & 0.04 & 0.17 & 0.23 & 0.29 & \cellcolor{gray!50}\textbf{0.39}\\
    H$\to$W 3 $\mapsto$ 7 & 1 & 0.25 & 0.32 & 0.34 &0.40 &  \cellcolor{gray!50}\textbf{0.42}\\
    H$\to$W 4 $\mapsto$ 17 & 1 & 0.31 & 0.39 & 0.41 & 0.40& \cellcolor{gray!50}\textbf{0.51}\\
    H$\to$W 5 $\mapsto$ 18 & 1 & 0.28 & 0.34 & 0.37 & 0.36 & \cellcolor{gray!50}\textbf{0.48}\\
    H$\to$W 6 $\mapsto$ 19 & 1 & 0.42 & 0.42 & 0.46 & 0.47& \cellcolor{gray!50}\textbf{0.49}\\
    H$\to$W 7 $\mapsto$ 20 & 1 & 0.39 & 0.41 & 0.41 & 0.44& \cellcolor{gray!50}\textbf{0.52}\\
    H$\to$W 8 $\mapsto$ 23 & 1 & 0.19 & 0.28 & 0.32 & 0.35& \cellcolor{gray!50}\textbf{0.46}\\
    \hdashline
    H$\to$W Avg & & 0.26 & 0.32 & 0.36& 0.39& \cellcolor{gray!50}\textbf{0.47}\\
    H$\to$W Std of Avg & & 0.05 & 0.05 & 0.03& 0.04& \cellcolor{gray!50}\textbf{0.03}\\
    \bottomrule
    \multicolumn{5}{l}{\scriptsize Higher H-Score is better. Best value in bold.}
  \end{tabular}
\end{table*}

\subsection{Ablation Studies}
\xhdr{Investigation of Loss Weights}
To account for the different magnitudes of the loss terms in \methodname, we employ weight balancing to ensure that the magnitudes of the loss terms are roughly comparable. We represent the overall loss as $L = \lambda_1 \cdot L_1 + \lambda_2 \cdot L_2 + \lambda_3 \cdot L_3$. The weights $\lambda_1$, $\lambda_2$, and $\lambda_3$ are normalized such that their sum is equal to 1:
$$
\lambda_1 = a / (a + b + c) ; \lambda_2 = b / (a + b + c) ; \lambda_3 = c / (a + b + c),
$$
where $a$, $b$, and $c$ are non-negative constants representing the desired relative importance of each loss term. To determine the optimal values of the weights $\lambda$ for each dataset, we perform a grid search using an independent source-target transfer scenario. Subsequently, we conduct experiments using the obtained weights across all transfer scenarios. The results of these experiments are presented in Table \ref{tab:weight}, which displays the average prediction accuracy for the target domains in the HAR dataset (closed-set DA). By employing weight balancing and optimizing the weights, we ensure that each loss term contributes appropriately to the overall objective of \methodname. This allows us to achieve better performance and more effective adaptation in various transfer scenarios. The results in Table \ref{tab:weight} demonstrate the impact of weight balancing and highlight the average prediction accuracy attained for the target domains.

\begin{table*}[h]
\centering
\caption{{Investigation on loss weights $a$, $b$, and $c$ for UniDA on WISDM using 1D-CNN as encoder.}}
\label{tab:weight}
\begin{tabular}{c|c|c|c}
\toprule
$a$ for Cross Entropy & $b$ for Sinkhorn & $c$ for Reconstruction & Accuracy \\\midrule
1 & 0.1 & 0.9 & 79.82 \\
1 & 0.2 & 0.8 & 80.54 \\
1 & 0.4 & 0.6 & 84.75 \\
1 & 0.6 & 0.4 & 86.37 \\
1 & 0.8 & 0.2 & 88.66 \\
1 & \textbf{1} & \textbf{0.2} & \textbf{94.26} \\
1 & 1 & 0 & 92.66 \\ \bottomrule
\end{tabular}
\end{table*}

\xhdr{Investigation of Sample Complexity}
We conducted an investigation into the impact of varying the amount of labeled source data on the performance of \methodname, along with several baseline methods. Specifically, we examined different proportions of labeled source data relative to the total source data (30\%, 50\%, 70\%, and 100\%) and evaluated the performance using prediction accuracy and F1 score on the target domain. The results of these experiments are presented in Table \ref{tab:sample}.The results demonstrate that \methodname consistently outperforms the baseline methods across all sample sizes. Even when only a limited amount of labeled source data is available, \methodname still achieves competitive performance, showcasing its robustness and effectiveness in scenarios with varying amounts of labeled source data. These findings provide valuable insights into the practical use of \methodname, particularly in real-world situations where obtaining labeled data can be challenging or resource-intensive. The ability of \methodname to leverage limited labeled source data and still achieve superior performance highlights its potential for practical applications and its capability to adapt well in settings where labeled data may be scarce.

\begin{table}[h]
\centering
\small
\caption{{Comparison of accuracy and F1 score on the HAR dataset for different domain adaptation methods with varying percentages of available source samples.}}
\label{tab:sample}
\begin{tabular}{c|cccc|cccc}
\multicolumn{1}{l|}{} & \multicolumn{4}{c|}{Accuracy} & \multicolumn{4}{c}{F1 Score} \\
\% of $\source$ & CDAN & DIRT-T & CLUDA & \methodname & CDAN & DIRT-T & CLUDA &\methodname \\ \midrule
30\% & 67.46$\pm$0.67 & 69.28$\pm$2.16 & 73.17$\pm$1.53 & \textbf{73.67$\pm$1.48} & 0.63$\pm$0.016 & 0.64$\pm$0.015 & 0.69$\pm$0.007 & \textbf{0.69$\pm$0.007}\\
50\% & 72.20$\pm$0.65 & 71.75$\pm$2.57 & 76.86$\pm$1.75 & \textbf{77.75$\pm$1.56} & 0.67$\pm$0.012 & 0.68$\pm$0.011 & 0.73$\pm$0.006 & \textbf{0.75$\pm$0.007}\\
70\% & 79.66$\pm$0.72 & 78.75$\pm$1.57 & 80.86$\pm$1.35 & \textbf{83.75$\pm$1.56} & 0.77$\pm$0.015 & 0.76$\pm$0.016 & 0.79$\pm$0.005 & \textbf{0.81$\pm$0.004}\\
100\% & 85.78$\pm$0.91 & 83.26$\pm$2.18 & 85.53$\pm$1.78 & \textbf{91.43$\pm$1.32} & 0.85$\pm$0.014 & 0.81$\pm$0.015 & 0.86$\pm$0.005 & \textbf{0.91$\pm$0.005} \\
\bottomrule
\end{tabular}
\end{table}

\section{Additional Discussion}
\label{sec:ablation}
 We first describe the importance of describing the application scenarios and the necessity of the task cannot be ignored. the goal of \methodname is to enhance the generalization of a machine learning model to an unlabeled target domain. The presence of feature and label shift between the source and target domains can lead to a decrease in model performance and accuracy. This emphasizes the need for Domain Adaptation techniques to improve the generalization and robustness of machine learning models in real-world scenarios. \methodname addresses both closed-set and universal domain adaptation, catering to different application scenarios and requirements. In closed-set domain adaptation, the focus is on adapting the model to a specific target domain while considering a fixed set of known classes or labels. On the other hand, universal domain adaptation expands the scope by handling the more challenging task of adapting to an unlabeled target domain that may contain unknown or novel classes. By addressing both closed-set and universal domain adaptation, \methodname provides a versatile framework that can be applied in a wide range of scenarios

\xhdr{Closed-Set Domain Adaptation (Closed-set DA)}
Closed-set DA is the problem of adapting a machine learning model trained on a labeled source domain to perform well on an unlabeled target domain where the set of classes is known in advance. Mitigating the feature shift is a common goal in this problem, where the distribution of features in the source domain differs from that in the target domain. Below are several examples of applications in different domains where Closed-Set Domain Adaptation is necessary:
\begin{itemize}
    \item Consider a time series classification task that aims to classify human activity based on accelerometer data. The distribution of features in accelerometer data collected during a weekday morning commute (source domain) may differ from that collected during a weekend hike (target domain). In this case, feature shift can occur due to changes in the distribution of features related to the user's movement patterns, such as walking speed, stride length, and acceleration profiles.
    \item Consider a speech recognition task; the acoustic features of speech signals may vary between different recording environments or speakers. For instance, a speech recognition model trained on speech data recorded in a quiet room (source domain) may have a different distribution of acoustic features when applied to speech data recorded in a noisy environment (target domain). In this case, feature shifts can occur due to changes in the distribution of features related to the background environment. 
\end{itemize}
\xhdr{Universal Domain Adaptation (UniDA)}
In real-world applications, little information may be available on the feature or label distribution of the target domain. Private labels in either the source or target domain may exist, i.e., classes present in one domain but absent in the other. This means feature and label shifts exist between source and target domains. Universal Domain Adaptation refers to the problem of adapting a machine learning model to perform well on a target domain under both feature and label shifts. UniDA allows machine learning models to generalize to new and diverse domains, improving their overall robustness and applicability in real-world scenarios. For example,
\begin{itemize}
    \item Consider a time series classification task to identify driving behaviors based on data collected from a car's sensors. The labels (e.g., aggressive driving, normal driving, or cautious driving) may vary between different drivers, depending on drivers’ driving style and data labeling methods. For example, the data from one driver (source domain) may record only aggressive and normal driving. In contrast, data from another driver (target domain) may record only normal and cautious driving due to differences in driving behaviors. In this case, the label shift can occur due to changes in the distribution of labels related to the driving habitats.
    \item Consider another time series EHR classification task where the goal is the prediction of hospital readmission. In the source domain, the labels could be defined as readmitted within 30 days, while in the target domain, the labels could be defined as readmitted within either 30 days or 60 days. This means the target domain has a different set of labels than the source domain, which could cause a label shift. In this case, the label shift could make it difficult for a machine learning model trained on the source domain to generalize well on the target domain.
\end{itemize}
In each of these examples, the feature and label shift between the source and target domains can decrease model performance and accuracy, highlighting the need for designing Domain Adaptation techniques to improve the generalization and robustness of machine learning models. Therefore, domain adaptation techniques like \methodname could be applied to align both feature and label shifts between the two domains and improve the model's generalization performance on the target domain. 

\subsection{The Use of Frequency Features}
It is important to consider the nature of the time series data when deciding whether frequency domain features are beneficial. In some cases, using frequency domain features may offer limited value, particularly when the data exhibits non-periodic or non-stationary patterns. For instance, in a time series dataset with a 2-way classification problem where both classes are driven by distinct temporal patterns at the same frequency rate, frequency features might not be as informative as time-based features. However, it is worth noting that \methodname is specifically designed to jointly model both time and frequency features, allowing the model to prioritize learning time features when frequency features are less informative. Thus, in \methodname, the potentially adverse effects of frequency features can be minimized due to the careful design of the time-frequency encoder.

It is essential to consider specific scenarios where the domain gap between the source and target domains is solely due to frequency changes. For example, in situations where the same data is collected using different experimental platforms that are calibrated differently in the source and target domains, methods that do not utilize frequency features may perform poorly. In \methodname, we adopt a simple approach by concatenating time and frequency features to ensure fair comparisons with baseline methods. However, it is worth exploring whether incorporating a transformer architecture could further improve the performance of \methodname over the existing time-frequency encoder. Transformers have demonstrated their effectiveness in capturing both local and global dependencies in sequential data, and their application to the time-frequency encoder in \methodname could potentially yield additional improvements.

To summarize, while frequency domain features may offer limited value in certain scenarios, \methodname is designed to handle such cases by allowing the model to prioritize time features when frequency features are less informative. Additionally, \methodname offers flexibility in adapting to scenarios where frequency changes contribute more to the feature shifts. The exploration of transformer architectures within \methodname presents an interesting direction for future research, as it may bring further improvements to the performance and adaptability of the model.

\subsection{Extension to Video Domain Adaptation}
In terms of interesting extensions, exploring the application of \methodname to video data can provide a more comprehensive evaluation and broaden its scope. Videos are more complex data types compared to simple time series, as they incorporate both spatial and temporal features. This complexity introduces additional challenges, such as varying visual styles, lighting conditions, and camera viewpoints, which can significantly impact the performance of machine learning models.

One relevant work in the field is the Temporal Attentive Adversarial Adaptation Network (TA3N) developed by Chen et al. \cite{Chen2019TemporalAA}. TA3N addresses video domain adaptation by simultaneously aligning and learning temporal dynamics without relying on sophisticated domain adaptation methods. It explicitly attends to temporal dynamics using domain discrepancy for effective domain alignment. Another notable framework is the unified framework for video domain adaptation presented by Kim et al. \cite{Kim2021LearningCC}, which focuses on regularizing cross-modal and cross-domain feature representations, as well as feature spaces.

To evaluate \methodname in the context of video domain adaptation, we conducted experiments on the publicly available benchmark dataset based on the UCF-HMDB benchmark, as assembled by Chen et al. \cite{Chen2019DomainAR}. This benchmark dataset consists of an overlapped subset of the original UCF and HMDB datasets, containing 3209 videos across 12 classes. We utilized the source code provided by the authors of TA3N \cite{Chen2019TemporalAA} and directly quoted the performance reported in their work. The results, shown in Table \ref{tab:videoDA}, highlight the effectiveness of \methodname on the UCF-HMDB dataset, providing promising outcomes and serving as a solid foundation for further research in video domain adaptation.

Exploring video domain adaptation within the framework of \methodname opens up new possibilities for addressing real-world challenges and enhancing the generalization and adaptability of machine learning models in video analysis tasks. This extension enables the consideration of both spatial and temporal features, contributing to more robust and accurate model performance in video domains. We plan to explore the efficacy of our \methodname on video domain adaptation in future work by considering efficient feature extraction through tensor decomposition and acceleration algorithms \cite{he2020fast, he2022gda,he2023meddiff, pmlr-v180-cai22a}. 

\begin{table}[]
    \centering
    \caption{The comparison of accuracy (\%) with other approaches on UCF-HMDB.}
\label{tab:videoDA}
\begin{tabular}{ c|c|c }
\toprule
Dataset & HDMB to UCF & UCF to HDMB \\
\midrule
DANN & 76.4 & 75.3 \\
TA3N & 81.8 & 78.3 \\
\methodname & 78.2 & 77.2 \\
\bottomrule
\end{tabular}
\end{table}

\subsection{Extension to Source-Free Domain Adaptation}
Source-free domain adaptation attracts increasing attention because, in many real-world scenarios, collecting labeled data from the source domain may be expensive, time-consuming, or even impossible \cite{Liu2021SourceFreeDA, Kundu2020UniversalSD,Yang2021GeneralizedSD, Xu2022SourceFreeVD}. In such cases, source-free domain adaptation allows leveraging a pre-trained model from a different source domain to adapt to a target domain to adapt without using labeled data from the source domain. It is a challenging and critical problem in machine learning, especially in computer vision tasks, where the source domain and target domain data have no overlap. SFDA aims to improve the performance of a model on a target domain with no access to any labeled data from the source domain. For example, \cite{Liu2021SourceFreeDA} proposed a method that leverages the structure of the image to learn domain-invariant features for the target domain via pixel-and patch-level optimization objectives tailored for semantic segmentation. Another approach to SFDA is generalized SFDA (G-SFDA) \cite{Yang2021GeneralizedSD}, which aims to handle the more challenging case where the target domain contains multiple domains. G-SFDA proposed a method using a structural clustering algorithm to group the target domain data into clusters based on their feature similarity. They then trained a model on each cluster to handle the domain shift. Universal source-free domain adaptation (USFDA) \cite{Kundu2020UniversalSD} is another variation of SFDA. It utilizes a novel instance-level weighting mechanism, source similarity metric (SSM), to handle both feature and label shifts. Recently, ATCoN \cite{Xu2022SourceFreeVD} is proposed to address Source-Free Video Domain Adaptation by learning temporal consistency, guaranteed by two novel consistency objectives, namely feature consistency and source prediction consistency, performed across local temporal features.

Indeed, one can extend \methodname to Source Free Domain Adaptation  (SF-DA) by only modifying the pre-training stage. During the pre-training stage of \methodname, the encoder $G_{\Time\Freq}$ is trained to learn well-separated, compact clusters of source domain data. This can be achieved by enforcing intra-class compactness and inter-class separability through negative classes, such as Triplet Loss. By doing so, the pre-trained model is better equipped for source-free deployment without prior knowledge of upcoming feature or label shifts.
Once a pre-trained model is obtained, it can be adapted to a target domain using the two-stage algorithm proposed in \methodname. In the first stage, the model encounters unlabeled target domain samples and obtains a target feature vector denoted as $\bz^t_{before}$. The correction step then updates the encoder $G_{\Time\Freq}$ and decoder $D_{\Time\Freq}$ by solving a reconstruction task on target samples, which repositions the target features $\bz^t_{before}$ into $\bz^t_{after}$. According to the cluster assumption that input data is separated into clusters with samples within the same cluster having the same label, the corrected encoder maintains the features of common target samples close to their originally assigned label while allowing the features of target unknown samples to diverge from their assigned label. \methodname can leverage this finding during deployment by detecting target unknown samples based on the movement of target features before and after the correction step. It assumes that if the distribution of the movement exhibits a bimodal structure, indicating the presence of unknown labels, it can easily detect private samples by training a 2-mean cluster while keeping common samples to their original assigned label.

\end{document}